\documentclass[10pt,journal,compsoc, draftcls,onecolumn]{IEEEtran} %
\ifCLASSOPTIONcompsoc
  \usepackage[nocompress]{cite}
\else
  \usepackage{cite}
\fi
 \pdfoutput=1
\usepackage{array}
\newcolumntype{L}[1]{>{\raggedright\let\newline\\\arraybackslash\hspace{0pt}}m{#1}}
\usepackage{amsmath}
\usepackage{amssymb}
\usepackage{verbatim}
\usepackage{multirow}
\usepackage{color,graphicx}
\usepackage{arydshln}
\providecommand{\zxhrefeq}[1]{(\ref{#1})}

\providecommand{\Zxhreftb}[1]{Table~\ref{#1}}
\providecommand{\zxhreftb}[1]{Table~\ref{#1}}
\providecommand{\zxhreffig}[1]{Fig.~\ref{#1}}
\providecommand{\Zxhreffig}[1]{Fig.~\ref{#1}}

\providecommand{\citep}[1]{\cite{#1}}

\begin{document}

\title{Multivariate mixture model for myocardial segmentation combining multi-source images}
\author{Xiahai~Zhuang
\IEEEcompsocitemizethanks{\IEEEcompsocthanksitem X. Zhuang is with School of Data Science, Fudan University, Shanghai, China.\protect\\
E-mail: zxh@fudan.edu.cn  \protect\\
Webpage: http://www.sdspeople.fudan.edu.cn/zhuangxiahai/}
\thanks{Manuscript received ..; revised ..}
}
\IEEEtitleabstractindextext{%
\begin{abstract}
This paper proposes a method for simultaneous segmentation of multi-source images, using the multivariate mixture model (MvMM) and maximum of log-likelihood (LL) framework.
The segmentation is a procedure of texture classification,
and the MvMM is used to model the joint intensity distribution of the images.
Specifically, the method is applied to the myocardial segmentation combining the complementary texture information from multi-sequence (MS) cardiac magnetic resonance (CMR) images.
Furthermore, there exist inter-image mis-registration and intra-image misalignment of slices in the MS CMR images. Hence, the MvMM is formulated with transformations, which are embedded into the LL framework and optimized simultaneously with the segmentation parameters.
The proposed method is able to correct the inter- and intra-image misalignment by registering each slice of the MS CMR to a virtual common space, as well as to delineate the indistinguishable boundaries of myocardium consisting of pathologies. Results have shown statistically significant improvement in the segmentation performance of the proposed method with respect to the conventional approaches which can solely segment each image separately. The proposed method has also demonstrated better robustness in the incongruent data, where some images may not fully cover the region of interest and the full coverage can only be reconstructed combining the images from multiple sources.
\end{abstract}

}

\maketitle
\IEEEdisplaynontitleabstractindextext
\IEEEpeerreviewmaketitle

\begin{figure*}[!tbhp]\center
\framebox{\includegraphics[width= 0.9\textwidth]{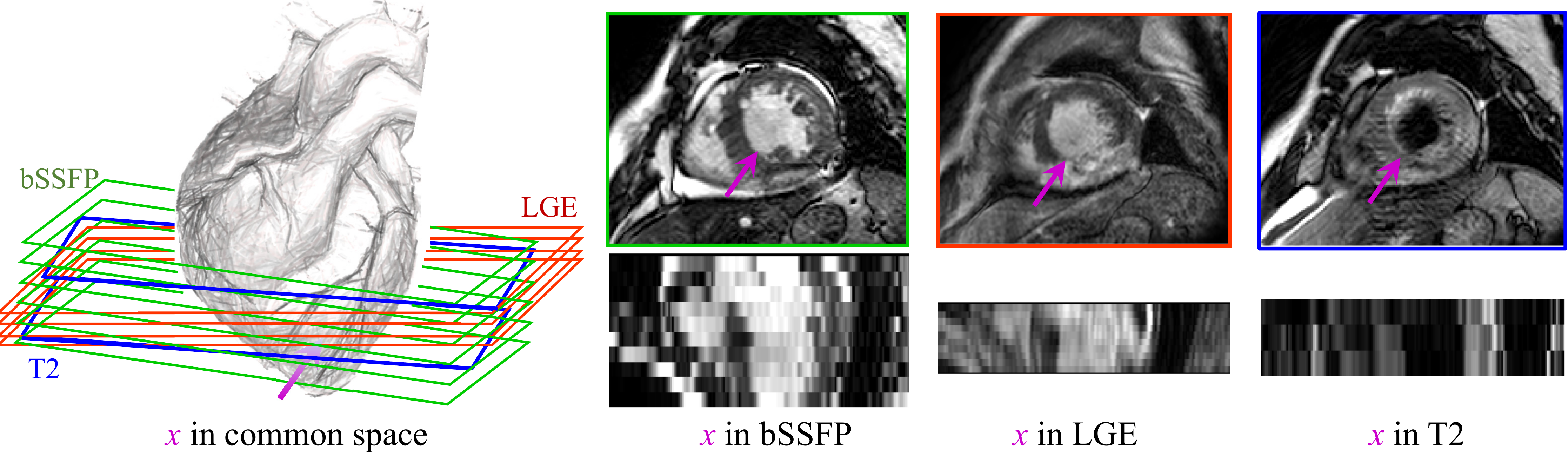}}
   \caption{Illustration of MVIs and the common space of a subject. Three CMR images are acquired from the patient, i.e. the bSSFP, LGE, and T2 CMR, which form the MVIs: $\boldsymbol{I}=\big[I_{bSSFP},I_{LGE},I_{T2}\big]$. The pink arrows indicate the same position at the common space and the three images which have different appearances and intensity values. }
\label{fig:commonspace}\end{figure*}

\IEEEraisesectionheading{\section{Introduction}\label{sec:introduction}}
\IEEEPARstart{S}{egmentation} from medical images is an essential prerequisite in a number of imaging data assisted medical applications, such as in the localization and quantification of tissues and pathologies, and modeling of anatomical structures \citep{journal/arbe/PhamXP00,journal/tmi/NobleB06,journal/mia/Petitjean11}.
Magnetic resonance (MR) technology provides an important tool for imaging anatomical and functional information of the heart, in particular the late gadolinium enhancement (LGE) cardiac magnetic resonance (CMR) sequence which visualizes myocardium infarction (MI), the T2-weighted CMR which images the acute injury and ischemic regions, and the balanced-Steady State Free Precession (bSSFP) cine sequence which captures cardiac motions and presents clear boundaries \citep{journal/circ/Kim09,journal/jacc/Kim09}.
\Zxhreffig{fig:commonspace} provides an example of the three sequences.
LGE CMR enhances the infarcted myocardium, to appear distinctive brightness compared with the healthy tissues, and therefore is effective in determining the presence, location, and extent of MI.
To identify the MI regions and subsequently perform the quantitative analysis, which is important in the diagnosis and treatment management of patients, a common method is to first delineate the myocardium from the LGE CMR images, where the pathologies can be differentiated from the healthy tissues using dedicatedly designed intensity-based threshold algorithms \citep{journal/ijci/Kolipaka05,journal/circ/Kim09,journal/jacc/Kim09,journal/jacc/Flett11}.



Since manual delineation is generally time-consuming, tedious and subject to inter- and intra-observer variations, automating this segmentation is desired in clinical practice, which is however still arduous, particularly due to the pathological myocardium from LGE CMR.
Besides the great variations of the heart shape across different subjects, there are three major issues related to the intensity distributions of the images, which challenge the conventional intensity-based classification methods, such as the Gaussian mixture model (GMM)-based segmentation \citep{journal/tmi/LeemputMS99,journal/mia/ValdesSR04}.
First, the intensity range of myocardium in LGE CMR overlaps its surroundings, leading to indistinguishable boundaries from its adjacent organs or regions.
For example, the infarcted myocardium, which is enhanced in LGE CMR, can appear identical to the blood pools, and the intensity range of the healthy myocardium can be similar to that of the adjacent liver or lung.
Also, the pathologies result in heterogeneous intensity of the myocardium, making the assumption of a simple distribution invalid; for example the single component Gaussian density function is widely used in the segmentation of bSSFP cine CMR \citep{journal/mia/ValdesSR04,journal/mia/Petitjean11}, but can not be used for LGE CMR.
Finally, the enhancement patterns can be complex. The location, size and shape of infarcted regions vary greatly across different patients, and the microvascular obstruction (MVO) sometimes occurs, which appears as hypo-enhanced areas (due to the lack of contrast agent uptake) within the hyper-enhanced regions. Hence, it is difficult to predict or make assumptions of the location and geometry of the MI regions.

\subsection{Related Work}

To the best of our knowledge, little work has been done in the fully automated myocardial segmentation from LGE CMR.
To obtain such segmentation, the algorithms generally need to integrate the prior shape information of the myocardium for guidance.
A number of reported methods used the segmentation result of the bSSFP cine CMR from the same subject, acquired in the same session, as \textit{a priori} knowledge.
The LGE CMR segmentation can then be achieved by directly propagating the bSSFP segmentation to the LGE CMR image space.
For this implementation, different registration methods can be used, such as 2D rigid registration based on a shift window \citep{conf/embc/Berbari09}, affine registration \citep{conf/miccai/Dikici04}, or rigid registration incorporating multi-scale total variation flow \citep{conf/isbi/Xu13}.
To detect the myocardium contours, a 2D geometrical template was proposed in \cite{conf/isbi/Ciofolo08}, where the myocardium was modeled as a closed ribbon structure with an imaginary centerline and varying width.
The myocardial template was divided into four quadrants, of which each was assigned an energy term based on the anatomical prior and knowledge of potential scars, for a deformable adaptation of the template to the target image.
In \citep{journal/tmi/Rajchl14}, a 3D mesh was built, based on the propagated prior segmentation from the bSSFP, and then deformed towards the myocardial contours in the LGE CMR to compensate for the difference between the slices from the two sequences.
In \cite{journal/mia/Wei13}, the authors proposed a 1D parametric model to detect the paired endocardial and epicardial edge points, where the intensity patterns along the radial rays from left ventricle (LV) center to beyond the epicardium were modeled. They imposed a thickness constraint for the 3D deformation.

In the literature, there is a research topic referred to as multivariate image analysis, which has been proposed to deal with images that have more than one measurement per pixel \cite{geladi1996multivariate,prats2011multivariate}.
This research has particularly focused on the three-channel (RGB) color images or multispectral and hyperspectral images,
and the methods have been designed to efficiently compress the highly correlated data and project them onto a reduced dimensional subspace, for example using the principal component analysis \cite{bharati1998multivariate}.
Multivariate image analysis considers only \emph{congruent} data, meaning for each pixel in one image there should be a corresponding pixel in the other images \cite{prats2011multivariate}.
However, in medical imaging the multi-source images commonly do not have this congruency, as the images acquired from different sources can be misaligned to each other, and the imaging field-of-views and resolution can vary greatly across different acquisitions.
The images having such differences are referred to as \emph{hetero-coverage} images in this work.

\subsection{Contribution of This Work}

The related works in the literature for myocardial segmentation from LGE CMR are mainly semi-automatic \citep{journal/tmi/Rajchl14}, or to propagate the segmented myocardium from one image as a constraint for the following separate segmentation of the other \citep{conf/miccai/Dikici04,journal/mia/Wei13}.
The separate segmentation however can be challenged to maintain a consistent and accurate segmentation result for the multi-source images or the multi-sequence (MS) CMR in this specific task of myocardial segmentation.
The consistency is important since the multi-source images come from the same subject,
and it can be obtained when the images are combined and segmented simultaneously in a unified framework.

This work proposes a unified framework which combines the complementary information from multi-source images and performs the segmentation of them \emph{simultaneously}.
This is achieved by formulating the segmentation of multi-source images, such as the MS CMR, using the multivariate mixture model (MvMM) and the maximum of log-likelihood (LL) framework.
The MvMM is adopted for modeling the joint intensity distribution of the multivariate images (MVIs), i.e. the multi-source images.

Furthermore, in MS CMR the image slices can be misaligned due to the respiratory motions of the subject during acquisition.
This misalignment not only happens between images from different sequences, but also happens among the slices from the same image.
It is worth mentioning that the atlas can be mis-registered to the target space.
The atlas is used to provide spatial constraints as well as to initialize the model parameters.
To correct these intra- and inter-image mis-registration, the MvMM is formulated with transformations, which are assigned to the atlas and each slice of the MS CMR images.
The transformation-embedded LL is then optimized using the iterative conditional mode (ICM) method, where
the MvMM parameters (segmentation) are obtained using the Expectation Maximization (EM) algorithm and the transformation (registration) parameters are updated using the gradient ascent optimization approach.

In this proposed framework, the combined segmentation can delineate the indistinct boundaries, such as these between MI regions and blood pools in LGE CMR, under the guidance of the T2 and bSSFP CMR, and vice versa. Also, the misalignment can be corrected using the combined information of the MVIs and the prior shape information contained in the atlas.

The rest of this paper is organized as follows:
Section~\ref{method} describes the proposed method in detail;
Section~\ref{result} presents the experimental setup and evaluation results,
using the clinical MS CMR images for assessing the myocardial segmentation and using the simulated MS brain MR for methodological study and comparisons, respectively.
In Section~\ref{conclude}, the author concludes this work and presents insights for potential extension of the proposed method.

\begin{figure*}[t]\center
 \framebox{\includegraphics[width=0.7\textwidth]{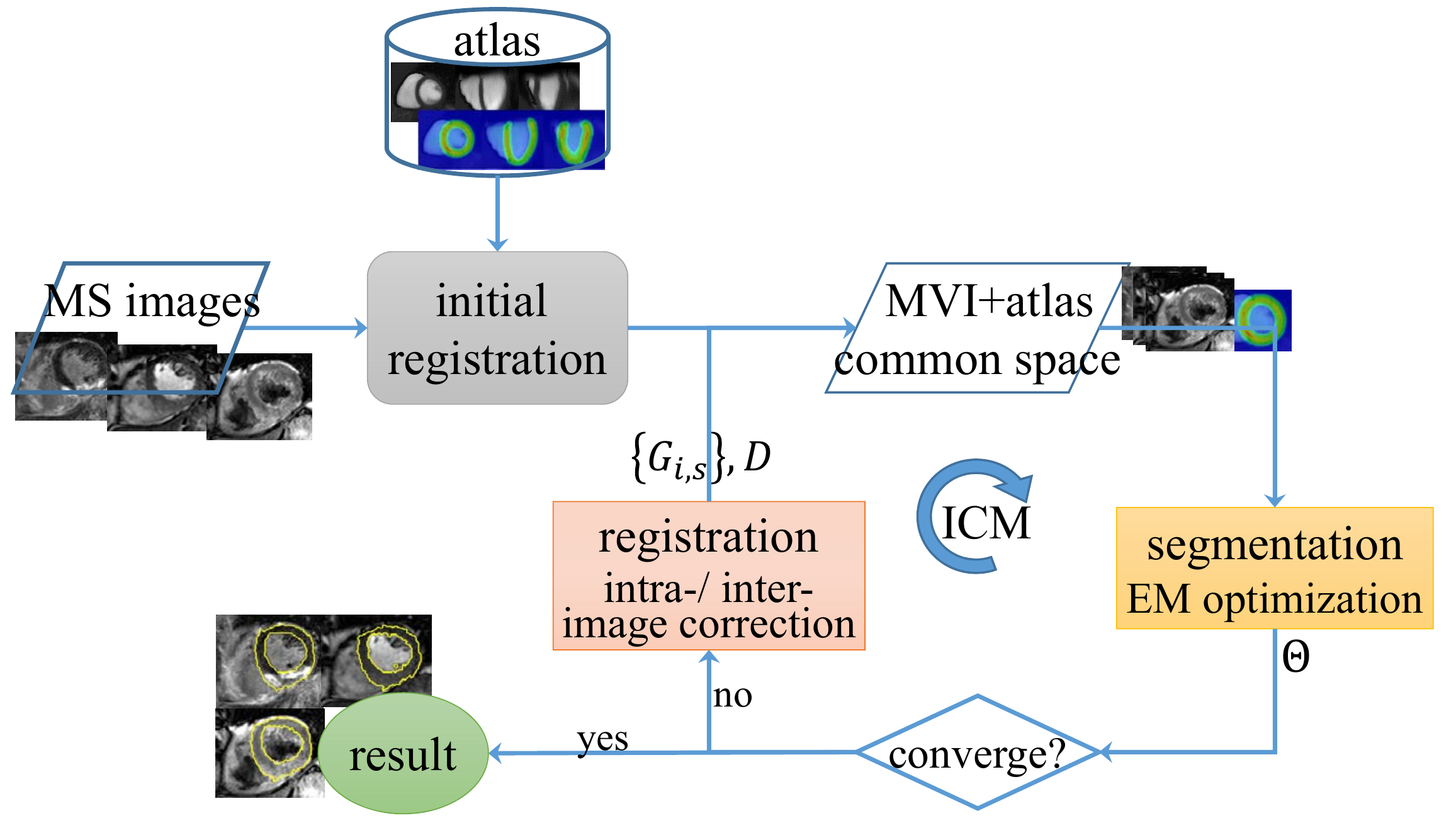}}
   \caption{Flowchart of the proposed myocardial segmentation method from the MS CMR.}
\label{fig:flowchart}\end{figure*}

\section{Method}\label{method}
The goal of this work is to classify the anatomical structures from the combined multi-source medical images acquired from the same subject. The images can be generated at different time points, using a variety of imaging modalities which provide diverse morphological and pathological information of the subject. By combining the multi-source information and performing the segmentation of the images simultaneously, one has the potential to segment the structures whose boundaries are not fully distinguishable in any one of the images.
To achieve this goal, the proposed method formulates the multi-source images using a MVI variable.

Let $\boldsymbol{I}\!\!=\!\!\{I_i|_{i=1...N_I}\}$ be the set of $N_I$ images acquired from the same subject, which form the MVIs.
One can denote the spatial domain of the region of interest (ROI) of the subject as $\Omega$, referred to as \emph{the common space}, which is the coordinate of the subject and thus defined by the combination of the MVIs in this formulation. \Zxhreffig{fig:commonspace} illustrates the concept of MVIs and the common space using the three CMR sequences.

For a location $x\!\!\in\!\!\Omega$, the tissue type of $x$ is determined regardless the appearance of the medical images.
One can denote tissue types using label values, namely $s(x)\!\!=\!\!k$, $k\!\!\in\!\! K$, $K$ is the set of labels,
and refer to subtypes of a tissue $k$ in image $I_i$ as $z_i(x)\!\!=\!\!c$, $ c\!\!\in\!\! C_{ik}$, $C_{ik}$ is the set of subtypes.
Note that provided the images are all aligned to the common space, the label information for all the MVIs should be the same, but the texture of images and classification of subtypes can be different.
For example, in brain MR images, the textures of brain tissues are different in the T1 and T2 weighted images; in CMR images, the scars of myocardial tissue are visible and distinguishable in the contrast LGE images, but not in the bSSFP or T2 images.

In the following, Section~\ref{method:mvmmll} introduces the MvMM and LL framework. Section~\ref{method:em} provides the EM approach and the initialization of the parameters.
The registration of MVIs is described in Section~\ref{method:reg},
and the optimization of the two sets of parameters is given in Section~\ref{method:icm}.
Finally, Section~\ref{method:inc} discusses and provides the implementation of MvMM when the MVIs have different coverage.
\Zxhreffig{fig:flowchart} presents the flowchart of the proposed segmentation framework.

\subsection{Multivariate Mixture Model in Log-Likelihood Framework} \label{method:mvmmll}

For a single image, one can use the mixture of Gaussian to model the intensity distributions, namely the GMM method, where the intensity probability density function (PDF) of one tissue is given by a Gaussian function. For a tissue with multiple subtypes, the multi-component GMM can be used \citep{conf/fimh/Shi11}.
For the MVI segmentation, one can use the MvMM.

The likelihood ($LH$) of the MvMM parameters $\theta$ is given by $LH_{\Omega}(\theta;\boldsymbol{I})\!\!=\!\!p(\boldsymbol{I}|\theta)$, similar to the GMM for classification, where $\boldsymbol{I}\!\!=\!\!\{I_1,\ldots I_{N_I}\}$ is the MVI vector. Assuming independence of each location (pixel), one gets $LH_{\Omega}(\theta;\boldsymbol{I})\!\!=\!\!\prod_{x\in\Omega}p(\boldsymbol{I}(x)|\theta)$.
The label at each location and component information at each image are hidden data.
\emph{Note that for the convenience and conciseness of denotation, we sometimes use the short terms $k_x$ and $c_{ix}$ (or $c_{ikx}$) to indicate  $s(x)\!\!=\!\!k$ and $z_i(x)\!\!=\!\!c$ (or $z_i(x)\!\!=\!\!c_{ik}$) when no confusion is caused.}

The conditional probability of MVIs at location $x$ given the model parameters can be computed as follows,
\begin{equation}
p(\boldsymbol{I}(x)|\theta)=\sum_{k\in K} \pi_{kx} p(\boldsymbol{I}(x)| s(x)\!\!=\!\!k,\theta) \ .
\label{eq:likehoodxcomplete}
\end{equation}
Here,
\begin{equation}
 \pi_{kx}=p(s(x)\!\!=\!\!k|\theta)=\frac{\pi_k p_A(k_x) }{N\!\!F}  ,
\label{eq:pikx}\end{equation}
where $p_A(k_x)$ is the atlas prior probability, $\pi_k$ is the label proportion, and $N\!\!F$ is the normalization factor.
When the tissue type of a position is known, \emph{the intensity distributions of different images become independent}, namely
\begin{equation}
p(\boldsymbol{I}(x)|s(x)\!\!=\!\!k,\theta)=\prod_{i=1,\ldots N_I}  p(I_i(x)| k_x,\theta) \ .
\label{eq:labellikelihood}
\end{equation}
The intensity PDF of an image, $p(I_i(x)| k_x,\theta)$, is given by the conventional multi-component GMM, as follows,
\begin{equation}
 p(I_i(x)| k_x,\theta) =\sum_{c\in C_{ik}}  \tau_{ikc} \Phi_{ikc}(\mu_{ikc},\sigma_{ikc},I_i(x)) \ ,
\label{eq:mcgmm}
\end{equation}
where, $\tau_{ikc}\!\!=\!\!p(c_{ikx} |k_x,\theta)$, s.t. $\sum_c \tau_{ikc}\!\!=\!\!1$, is the component proportion, and $\Phi_{ikc}(\cdot)\!\!=\!\!p(I_i(x)|c_{ikx},\theta)$ is the Gaussian function modeling the intensity PDF of a tissue subtype $c$ of tissue $k$ in the image $I_i$.

\subsection{Expectation Maximization Algorithm}\label{method:em}

The LL of the complete data, which includes both the observed MVIs and the hidden data, is as follows,
\begin{equation}
\begin{array}{r@{}c@{}l}
 LL_{com} &=& \displaystyle\sum_{x\in\Omega}\sum_{k\in K} \delta_{s(x),k}\Big\{ \log\pi_{kx}    \\
     &&+ \displaystyle\sum_i\sum_{c_{ik}} \delta_{z_i(x),c_{ik}}\big(\log \tau_{ikc}+\log \Phi_{ikc}(I_i(x))\big) \Big\} \\
    &=& \displaystyle\sum_x\sum_k \delta_{s(x),k} \log\pi_{kx}    \\
     &&+ \displaystyle\sum_x\sum_k\sum_i\sum_{c_{ik}} \delta_{s(x),k}\delta_{z_i(x),c_{ik}} \texttt{Term}_{\texttt{LTP}} ,
     \end{array}
\label{eq:ll}
\end{equation}
where $\delta_{a,b}$ is the Kronecker delta function, and
$\texttt{Term}_{\texttt{LTP}} \!=\!(\log \tau_{ikc}\!\!+\!\!\log \Phi_{ikc}(I_i(x)))$.
The conditional expectation of $LL_{com}$ given the current parameter $\theta^{[m]}$ at the $m$-th step and the observed MVIs is  $Q(\theta|\theta^{[m]})\!\!=\!\!E(LL_{com}|\boldsymbol{I},\theta^{[m]})$.
Therefore, one can employ the following E-step and M-step iteratively to obtain the model parameters and then the segmentation variables.

 \textbf{E-Step}:
One can obtain the expectation of $LL_{com}$ by computing the expectation of $\delta_{s(x),k}$ and $\delta_{s(x),k}\delta_{z_i(x),c_{ik}}$ given the observed MVIs and current estimation of model parameters,
\begin{equation}   \begin{array}{l @{ } l}
P^{[m+1]}_{kx}&=E_{(\boldsymbol{I},\theta^{[m]})}\left(\delta_{s(x),k}\right)\\
  &= p\big(s(x)\!\!=\!\!k|\boldsymbol{I},\theta^{[m]}\big)\\
  &=\displaystyle\frac{p(\boldsymbol{I}(x)|k_x,\theta^{[m]})\pi^{[m]}_{kx}}{\sum_{l\in K} p(\boldsymbol{I}(x)|l_x,\theta^{[m]}) \pi^{[m]}_{lx} } \ ,
\end{array}\label{eq:esteppm}\end{equation}
and,
\begin{equation}   \begin{array}{l @{}l}
P^{[m+1]}_{ikcx}&=E_{(I_i,\theta^{[m]})}\left(\delta_{s(x),k}\delta_{z_i(x),c_{ik}}\right)  \\
  &= p\left(s(x)\!\!=\!\!k,z_i (x)\!\!=\!\!c_{ik}|\boldsymbol{I},\theta^{[m]} \right) \\
  &= p\big( c_{ikx}| k_x, \boldsymbol{I},\theta^{[m]}\big) P^{[m+1]}_{kx} \\
  &=\displaystyle\frac{ p(\boldsymbol{I}(x)|c_{ikx},k_x,\theta^{[m]}) \tau_{ikc}^{[m]}} {p(\boldsymbol{I}(x)| k_x,\theta^{[m]})} P^{[m+1]}_{kx}\\
  &=\displaystyle\frac{ \Phi(\mu_{ikc}^{[m]},\sigma_{ikc}^{[m]},I_i (x)) \tau_{ikc}^{[m]}} {p(I_i(x)| k_x,\theta^{[m]})} P^{[m+1]}_{kx}
  \ ,
\end{array}\label{eq:estep}
\end{equation}
where,
$p(\boldsymbol{I}(x)|c_{ikx},k_x,\theta^{[m]})$ is computed using $p(I_i(x)|c_{ikx},\theta^{[m]})\prod_{j\neq i}p(I_j(x)|k_x,\theta^{[m]})$,
\emph{which is based on the same assumption for $p(\boldsymbol{I}(x)| k_x,\theta^{[m]})$ in \zxhrefeq{eq:labellikelihood} that the intensity distributions of MVIs becomes independent when the label and component information are known}, and the common part, $\prod_{j\neq i}p(I_j(x)|k_x,\theta^{[m]})$, in the numerator and denominator is then cancelled out;
$\pi^{[m]}_{kx}$ is computed using \zxhrefeq{eq:pikx}, where $\pi^{[m]}_{k}$ is calculated in the previous M-step.
Here, $P^{[m+1]}_{ikcx}$ and $P^{[m+1]}_{ikcx}$ are also referred to as the estimation results of the hidden data.

\textbf{M-Step}:  The model parameters are updated by analytically maximizing $Q(\theta|\theta^{[m]})$,
\begin{equation} \begin{array}{r @{}l}
\tau_{ikc}^{[m+1]}&=\frac{\sum_{x\in\Omega} P_{ikcx}^{[m+1]}}{\sum_{d\in C_{ik}} \sum_{x\in\Omega} P_{ikdx}^{[m+1]}} \\
\mu_{ikc}^{[m+1]}&=\frac{\sum_{x\in\Omega} I_i(x) P_{ikcx}^{[m+1]}}{\sum_{x\in\Omega} P_{ikcx}^{[m+1]} }\\
(\sigma_{ikc}^{[m+1]})^2 &= \frac{\sum_{x\in\Omega} (I_i(x)-\mu_{ikc}^{[m+1]} )^2 P_{ikcx}^{[m+1]}}
{\sum_{x\in\Omega} P_{ikcx}^{[m+1]}} \ .
\end{array}
\label{eq:mstep}\end{equation}
For $\pi^{[m+1]}_k $,  one can compute
\begin{equation} \begin{array}{r@{\ }l}
\displaystyle\frac{\partial Q(\theta|\theta^{[m]})}{\partial \pi_k}
    =& \displaystyle\sum_x \frac{P^{[m+1]}_{kx}}{\pi_k} - \sum_x\sum_{j\in K}\frac{P^{[m+1]}_{jx}p_A(s(x)\!\!=\!\!k)}{\sum_{l\in K}p_A(s(x)\!\!=\!\!l)\pi_l}\\
    =& 0  \ . \end{array}
\label{eq:pik}\end{equation}
Since there is no closed form for the solution, one can use numerical methods such as the iterative gradient ascent method to search the optimal solution for $\pi^{[m+1]}_k$. Alteratively, one can regard $\mathcal{C}_x^{[m]}\!\!=\!\!\sum_{l\in K}p_A(l_x)\pi_l$ as a constant using $\pi_l\!\!=\!\!\pi^{[m]}_l$ in the [m+1]th iteration, which results in
\begin{equation}
\pi^{[m+1]}_k=\frac{\sum_x P^{[m+1]}_{kx}}{
\sum_x \left({p_A(k_x)/ \mathcal{C}_x^{[m]}} \right)
} .
\label{eq:pim1k}\end{equation}
This guarantees an improvement of the likelihood of the parameters with respect to the MVIs in each iteration and leads to a generalized EM scheme \citep{journal/ni/Ashburner05}.

\textbf{Initialization} of $\pi_{k}^{[0]}$, $\tau_{ikc}^{[0]}$, $\mu_{ikc}^{[0]}$ and $\sigma_{ikc}^{[0]}$ are computed based on the atlas prior probabilities which are propagated using atlas-to-target image registration technology \citep{journal/tmi/Zhuang10},
\begin{equation} \begin{array}{r@{\ }l@{\ }l}
 \pi_k^{[0]}=&\displaystyle\frac{\sum_x p_A(k_x)}{\sum_{l\in K}\sum_x p_A(l_x)},&\tau^{[0]}_{ikc}=\frac{1}{|C_{ik}|} \\
 \mu^{[0]}_{ikc}=&\left\{ \begin{array}{ll}
       \mu^{[0]}_{ik} + a\sigma^{[0]}_{ik}, &|C_{ik}\geq 2| \\
       \mu^{[0]}_{ik}, &|C_{ik}=1| \\
  \end{array}
  \right. , & \left(\sigma^{[0]}_{ikc}\right)^2=\frac{\sigma^{[0]}_{ik}}{|C_{ik}|} ,
 \end{array}
\label{eq:init}\end{equation}
where $a$ is a proportion factor to uniformly sample $|C_{ik}|$ values from $[-1,1]$ for the $|C_{ik}|$ components of label $k$ in image $I_i$; $\mu^{[0]}_{ik}\!\!=\!\!\frac{\sum_x I_i(x)p_A(k_x)}{\sum_x p_A(k_x)}$, and $\sigma^{[0]}_{ik}\!\!=\!\!\frac{\sum_x (I_i(x)-\mu^{[0]}_{ik})^2p_A(k_x)}{\sum_x p_A(k_x)}$.

\begin{figure*}[tb]\center
 \includegraphics[width= \textwidth]{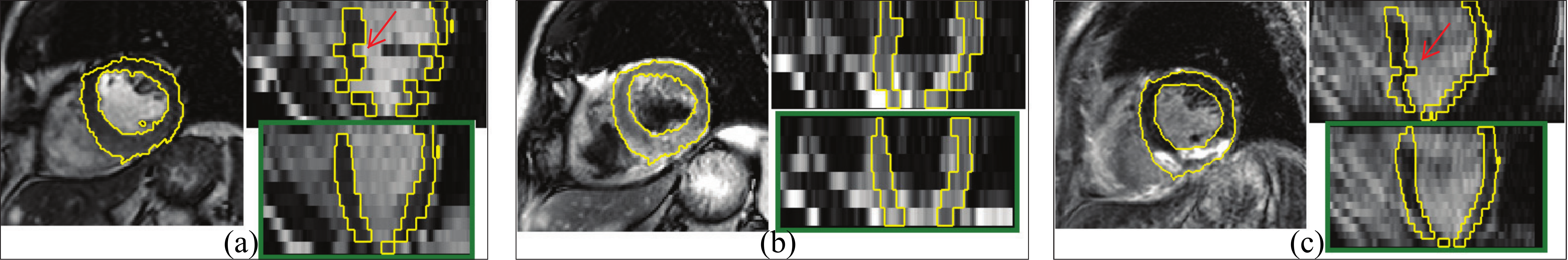}
   \caption{Segmentation results of the three CMR sequences, (a) bSSFP, (b) T2-weight, (c) LGE; myocardial boundaries are highlighted in yellow color; the motion shifts are pointed out by the red arrows.
The images in the green boxes of (a)-(c) are the shift corrected images.}
\label{fig:misalignment}\end{figure*}

\subsection{Registration in Multivariate Mixture Model} \label{method:reg}
The MvMM is initialized and regularized by the prior probabilities from an atlas which can be registered to the target MVIs using the conventional methods in the atlas-based segmentation framework  \citep{journal/mia/Zhuang16}.

However, there exist two types of misalignment.
Firstly, the motion shift of each slice is commonly seen in the multi-slice CMR, besides the misalignment of the whole image to the common space of the subject.
\Zxhreffig{fig:misalignment} visualizes the motion shifts in the \textit{in vivo} CMR and the corrected images.
Secondly, the atlas, providing the prior probabilities, can be mis-registered to the common space in some local details.

The motion shift of a slice can be modeled by an affine transformation, leading to reformulation of the intensity density of a subtype tissue as follows,
\begin{equation}
 p(I_i(x)|c_{ik},\theta, \{G_{i,s}\})=\Phi_{ikc}(I_i(G_{i,s}(x))) ,
\end{equation}
where $\{G_{i,s}\}$ are the affine transformations for all slices. The atlas deformation, denoted using $D$, can be embedded into the prior probabilities for correcting the local mis-registration,
\begin{equation}
 p_A(s(x)\!\!=\!\!k|D) \!= \!p_A(s(D(x))\!\!=\!\!k) \!=\! A_k(D(x)), k\!\!=\!\!1\ldots K \ ,
\end{equation}
which are the probabilistic atlas images. The original LL then becomes,
\begin{equation} \begin{array}{ @{}l@{}}
  LL(\theta,D,\{G_{i,s}\})\\
 \quad  = \displaystyle\sum_{x\in{\Omega}} \log
        \Big\{ \sum_k p(s(x)\!\!=\!\!k|D)
             \prod_i \sum_{c_{ik}} \tau_{ikc}\Phi_{ikc}(I_i(G_{i,s}(x))) \Big\} \\
 \quad = \displaystyle\sum_{x\in{\Omega}} \log LH(x) . \end{array}
\label{eq:mvill}\end{equation}
Here, the prior is define as $p(s(x)\!\!=\!\!k|D)\!\!=\!\!A_k(D(x)) \pi_{kx}/\mathcal{N}$, where $\mathcal{N}$ is the normalization factor\citep{journal/mia/ValdesSR04}.

\subsection{Iterative Conditional Mode Optimization}\label{method:icm}

There is no closed form solution for minimization of \zxhrefeq{eq:mvill}.
Since the Gaussian parameters depend on the values of the transformation parameters, and vice versa, one can use the ICM approach to solve this optimization problem, which is eventual a coordinate ascent method in this formulation \citep{conf/spie/Lee00}.
The ICM scheme optimizes one group of parameters while keeping the others unchanged at each iteration. The different groups of parameters are alternately optimized and this alternation process iterates until a local optimum is found.
\Zxhreffig{fig:flowchart} provides the flowchart of the framework, where the segmentation parameters $\Theta$, i.e. the MvMM parameters and the hidden data, are updated using the EM approach (Section~\ref{method:em}), and the registration parameters, i.e. the transformations $\{G_{i,s}\}$ and $D$, are optimized using the gradient ascent method.
The derivatives of LL with respect to the affine transformations and atlas deformation are respectively given by,
\begin{displaymath}\begin{array}{r@{ }r}
\displaystyle\frac{\partial LL}{\partial G_{i,s}} &= \displaystyle\sum_x \frac{1}{LH(x)}\sum_k  p(s(x)\!\!=\!\!k|D) \prod_{j\neq i} \Big\{ p(I_j(x)|s(x)\!\!=\!\!k) \\
               &   \cdot \sum_c(\tau_{ikc}\Phi'_{ikc}\nabla I_i(y)\nabla G_{i,s}(x)) \Big\}
\end{array}
\end{displaymath}
where $y\!\!=\!\!G_{i,s}(x)$, and
\begin{equation}
\displaystyle\frac{\partial LL}{\partial D} = \displaystyle\sum_x \frac{1}{LH(x)}\sum_k  \frac{\partial p(s(x)\!\!=\!\!k|D)}{\partial D}
 p(\boldsymbol{I}(x)|s(x)\!\!=\!\!k),
\end{equation}
where the computation of $\frac{\partial p(s(x) = k|D)}{\partial D}$ is related to $\frac{\partial A_k(D(x))}{\partial D}$, and
\begin{equation}
 \frac{\partial A_k(D(x))}{\partial D} = \nabla A_k|_{y=[y_1,y_2,y_3]} \times \left[ \frac{\partial y_1}{\phi_d},\frac{\partial y_2}{\phi_d},\frac{\partial y_3}{\phi_d} \right]^{\texttt{T}}.
\end{equation}
Here, $y\!\!=\!\!D(x)$, and $\{\phi_d\}$ are the free-form deformation parameters \citep{journal/tmi/Rueckert99,journal/tmi/Zhuang11}.

\subsection{Hetero-Coverage Multivariate Images} \label{method:inc}

\begin{figure*}[thb]\center
 \framebox{\includegraphics[width=0.6\textwidth]{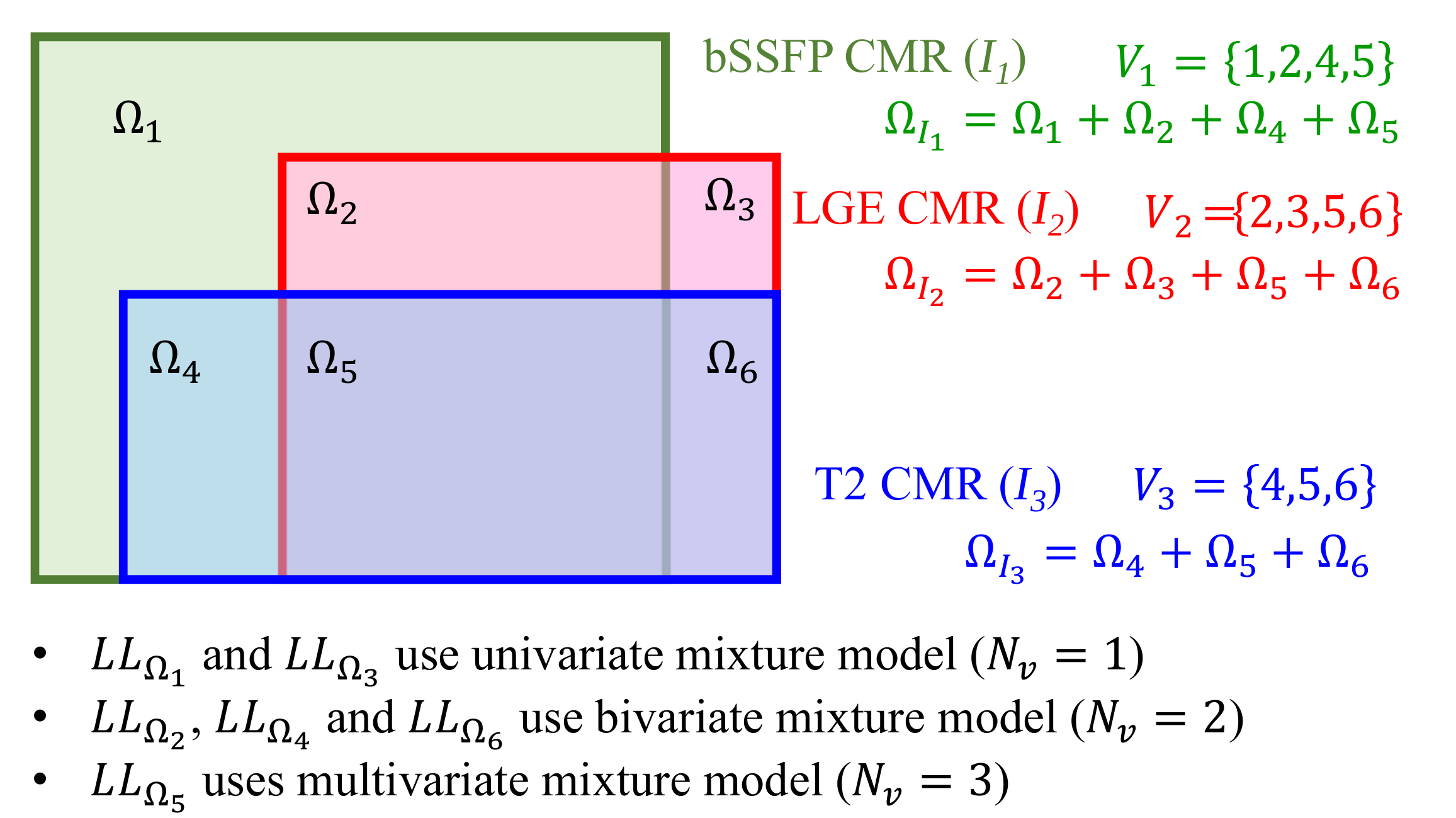}}
   \caption{Illustration of MVIs with different coverage, where the log-likelihood $LL_{\texttt{HC}}\!\!=\!\!\sum_{v=1}^{6} \!\!LL_{\Omega_v}$.}
\label{fig:inc}\end{figure*}

In medical imaging, data from different acquisitions can generally have different resolutions and coverage of the ROI. The combination of these images is referred to as hetero-coverage multivariate images (HC-MVIs).
For example, the bSSFP CMR sequence generally covers the whole ventricle, while the LGE and T2 sequences may only acquire the data from major part of the main ventricles without covering the apex or base, as \zxhreffig{fig:commonspace} shows.
Also, the missing information between the slices may be inappropriate to obtain by image interpolation techniques due to the large thickness of slices and the gap between them, i.e. low resolution in the inter-slice dimension.

This section generalizes the formulation of MvMM for such HC-MVIs where some content is not acquired in certain images.
In the new formulation, the ROI of the subject is divided into $N_{sr}$ non-overlapping sub-regions, $\{\Omega_v|_{v=1\ldots N_{sr}}\}$, and $\Omega\!\!=\!\!\bigcup_{v=1}^{N_{sr}} \Omega_v$.
For image $I_i$, the volume is denoted as $\Omega_{I_i}$, and the set of indices of sub-regions belonging to $\Omega_{I_i}$ is denoted as $V_i$.
Therefore, $\Omega_{I_i}$ is the sum of the set of sub-regions belonging to $I_i$, i.e., $\Omega_{I_i} \!\!=\!\! \bigcup_{v\in V_i} \Omega_{v}$.
Also, each sub-region $\Omega_v$ fully covers a sub-volume of $N_v$ images, $N_v\!\!\in\!\![1,N_I]$,  and thus one can apply a conventional $N_v$-variate mixture model for $\Omega_v$.

\Zxhreffig{fig:inc} illuminates the sub-regions of HC-MVIs, and as it shows, one can rewrite the LL as follows,
\begin{equation}
LL_{\texttt{HC}}=\sum_{v=1}^{N_{sr}} LL_{\Omega_v} =\sum_{v=1}^{N_{sr}} \log LH_{\Omega_v}(\theta; \boldsymbol{I}_v),
\label{eq:inc:ll}\end{equation}
where,
\begin{equation}
LL_{\Omega_v}= \sum_{\Omega_v} \log \sum_{k\in K} \Big\{ \pi_{kx}\!\! \prod_{i=1\ldots N_v}\!\!\Big(\! \sum_{c\in C_{ik}}\!\tau_{ikc} \Phi_{ikc} (I_i (x) ) \Big) \Big\} \ . \label{eq:inc:llv}\end{equation}
Here, $\boldsymbol{I}_v$ could be one image, such as solely for the bSSFP image when neither of the T2 or LGE image covers sub-region $\Omega_v$, which results in a conventional univariate GMM.

Since the $LL_{\texttt{HC}}$ is the sum of the LL terms from a set of standard MVI models, the optimization of the parameters is similar to that of LL in Section~\ref{method:em}.

\textbf{E-Step:}
For $x\!\!\in\!\!\Omega_v$ and the MVIs in $\Omega_v$ is $\boldsymbol{I}_v$, the estimation of the hidden label in \zxhrefeq{eq:esteppm} is computed as follows,
\begin{equation} \begin{array}{r@{\ }l}
P_{kx}^{[m+1]}= & p(s(x)\!\!=\!\!k|\boldsymbol{I}_v,\theta^{[m]} ) \\
= & \displaystyle\frac{p(\boldsymbol{I}_v(x)|k_x,\theta^{[m]})\pi^{[m]}_{kx}}{\sum_{l\in K} p(\boldsymbol{I}_v(x)|l_x,\theta^{[m]}) \pi^{[m]}_{lx} } \end{array} ,
\label{eq:inc:em:pkx}\end{equation}
where $p\big(\boldsymbol{I}_v(x)|k_x,\theta^{[m]}\big)\!\!=\!\!\prod_{i=1,\ldots N_v}  p(I_i(x)| k_x,\theta^{[m]})$ is computed following \zxhrefeq{eq:labellikelihood} and  $\pi^{[m]}_{kx}\!\!=\!\!\frac{\pi_k^{[m]}  p_A(k_x) }{N\!\!F}$.
Having $P_{kx}^{[m+1]}$ for $\forall x\!\!\in\!\!\Omega$, one can obtain $P_{ikcx}^{[m+1]}\!\!=\!\!\frac{ \Phi(\mu_{ikc}^{[m]},\sigma_{ikc}^{[m]},I_i (x)) \tau_{ikc}^{[m]}} {p(I_i(x)| k_x,\theta^{[m]})} P^{[m+1]}_{kx}$ for $\forall x\!\!\in\!\!\Omega_{I_i}$, similar to \zxhrefeq{eq:estep}.

\textbf{M-Step:}
By maximizing the $Q$ function, one obtains,
\begin{equation} \begin{array}{r @{}l}
\tau_{ikc}^{[m+1]}
    &=\displaystyle\frac{ \sum_{v\in V_i}\sum_{x\in\Omega_v} P_{ikcx}^{[m+1]}}{\sum_{d\in C_{ik}} \sum_{v\in V_i}\sum_{x\in\Omega_v} P_{ikdx}^{[m+1]}} \\
    &=\frac{ \sum_{x\in\Omega_{I_i}} P_{ikcx}^{[m+1]}}{\sum_{d\in C_{ik}} \sum_{x\in\Omega_{I_i}} P_{ikdx}^{[m+1]}} \\
\mu_{ikc}^{[m+1]}&=\displaystyle\frac{\sum_{x\in\Omega_{I_i}} I_i(x) P_{ikcx}^{[m+1]}}{\sum_{x\in\Omega_{I_i}} P_{ikcx}^{[m+1]} }\\
(\sigma_{ikc}^{[m+1]})^2 &= \displaystyle\frac{\sum_{x\in\Omega_{I_i}} (I_i(x)-\mu_{ikc}^{[m+1]} )^2 P_{ikcx}^{[m+1]}}
{\sum_{x\in\Omega_{I_i}} P_{ikcx}^{[m+1]}} \ .
\end{array}
\label{eq:inc:mstep}\end{equation}

For $\pi_k^{[m+1]}$, its computation is solely related to the atlas prior probability and the estimation of the hidden data, which are all obtained within the common space using the different multi-variate image models. Therefore, one can compute $\pi_k^{[m+1]}$ using the same updating formula in \zxhrefeq{eq:pik} and \zxhrefeq{eq:pim1k}.

Initialization of the parameters and optimization of the registration parameters are similar to the computation for the congruent data in Section~\ref{method:em} and ~\ref{method:reg}, except that they are based on the sum of the sub-regions $\{\Omega_v|_{v\in V_i}\}$ for the variables of image $I_i$.

\section{Experiments and Results}\label{result}

This section consists of three experiments to evaluate the proposed MvMM segmentation method.
Section~\ref{exp:mvmm} evaluates the performance of the proposed MvMM for myocardial segmentation from MS CMR.
The information of the CMR data, evaluation metrics and implementation details are presented.
The inter-observer and inter-sequence variations are also studied here.
Section~\ref{exp:alternative} compares the performance of different segmentation methods, including the conventional approaches and the MvMM with alternative registration schemes.
The study focuses on the segmentation of LGE CMR, because it is challenging and the main focus of the research.
Section~\ref{exp:brain} investigates the performance of the proposed MvMM with different dimensions of the multivariate variable and HC-MVIs.
This study uses the simulated brain images from Brain Web \cite{cocosco1997brainweb}.

\subsection{Segmentation Combining Multi-Sequence CMR}\label{exp:mvmm}

\begin{table*}[tb]\center
\caption{Information of the CMR data.}\label{tb:imageinfo}
\begin{tabular}{l c c c c }
\hline
CMR    & slice pixel size & slice thickness & no. of slices & coverage \\ \hline
LGE     & $0.75\times0.75$ mm & 5 mm & 10 - 18 & main body of ventricles\\
T2     & $1.35\times1.35$ mm & 12-20 mm & 3 - 7 & main body of ventricles\\
bSSFP  & $1.25\times1.25$ mm & 8-13 mm & 8 - 12 & full ventricles\\ \hline
\multicolumn{3}{l}{Number of patients: \quad 35}	 \\\hline
\end{tabular}\end{table*}

\subsubsection{Data}
\textbf{CMR:} The CMR data from 35 patients, who underwent cardiomyopathy, had been collected from Shanghai Renji hospital with institutional ethics approval and had been anonymized.
Each patient had been scanned using the three CMR sequences, i.e. the LGE, T2 and bSSFP, from short-axis orientation.
The LGE CMR images consist of more than ten slices, covering the main body of the ventricles, and they were acquired and reconstructed into resolution about $0.75\times 0.75$ mm in-plane and 5 mm slice thickness.
The T2 CMR images only have a few slices: 13 cases having three slices, and the others having five (13 subjects), six (8 subjects) or seven (one subject) slices.
They were acquired and reconstructed into resolution about $1.35\times1.35$ mm in-plane and 12 to 20 mm slice thickness.
The bSSFP images are cine data and fully cover the ventricles from the apex to the basal plane of the mitral valve, with some cases having several slices beyond the ventricles. The images were acquired and reconstructed into resolution about $1.25\times1.25$ mm in-plane and 8 to 13 mm slice thickness.
Since both the LGE and T2 CMR were scanned at the end-diastolic phase, the same cardiac phase of the bSSFP cine images were identified. Therefore, the bSSFP CMR refers to the single phase data in this work.
The detailed information and the parameters of the three CMR sequences are summarized in \zxhreftb{tb:imageinfo}.

\textbf{Gold standard:} Each of the CMR images had been manually delineated three times, by three independent, well-trained observers who were not aware of the methodology of this work.
The manual segmentation was performed slice-by-slice, using the brush tool in the ITK-SNAP \citep{itksnap},
and the gold standard segmentation was the average of the three manual delineations.

\begin{figure*}[tb]\center
    \begin{tabular}{cc}
     \includegraphics[width=0.48\textwidth]{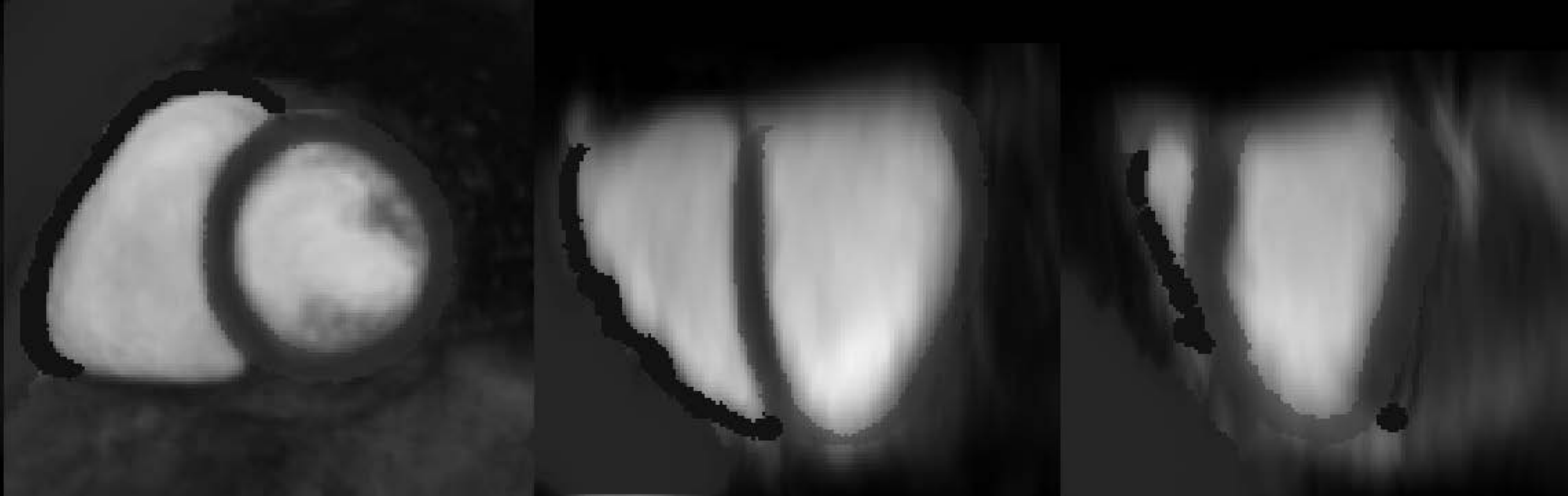} & \includegraphics[width=0.48\textwidth]{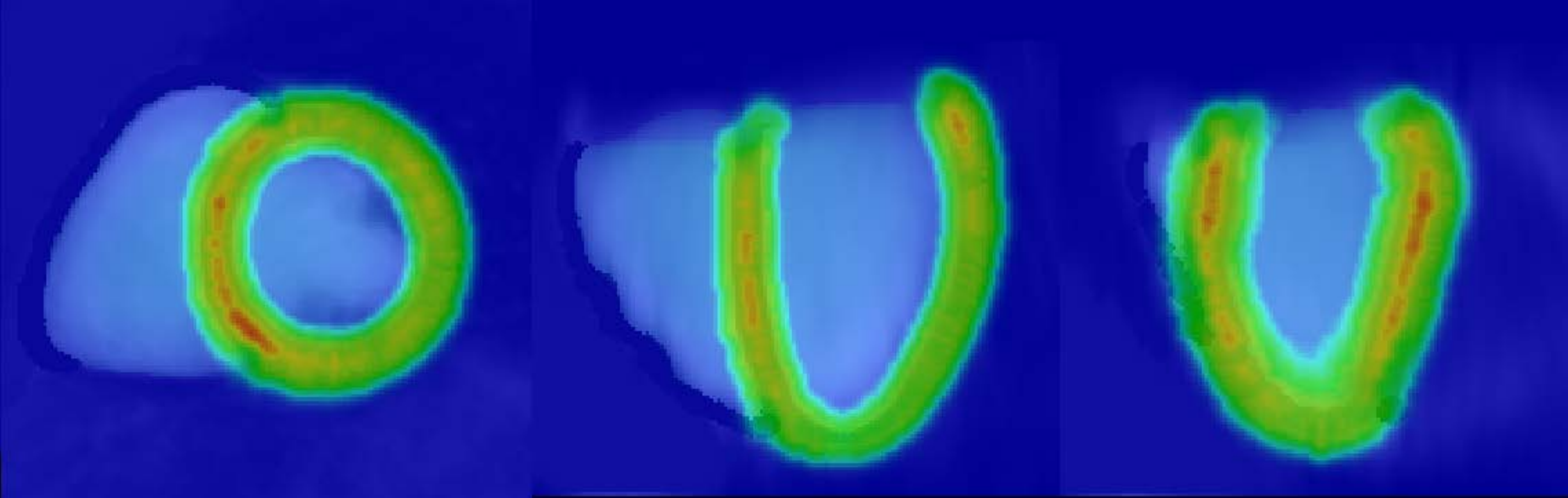} \\
    (a) & (b)\end{tabular}
    \caption{The three orthogonal views of the atlas intensity image (a), and the corresponding probabilistic atlas of myocardium superimposed on the intensity image (b).}
\label{fig:atlas}  \end{figure*}

\textbf{Atlas:} The atlas was built from another set of bSSFP images from healthy subjects. The bSSFP images were first manually segmented, and then nonrigidly registered to a selected reference using a comprehensive registration method \cite{journal/tmi/Zhuang10}. The atlas intensity image and label probabilities can be computed based on the mean of all the transformed images and label information \citep{journal/mia/ValdesSR04}.
To enhance the right ventricular (RV) boundary of the atlas intensity image which were blurred due to the intensity averaging, we manually added a low-intensity region to the atlas image, to simulate an intensity-distinct epicardial boundary of the RV.
To obtain a more uniform and less biased probability map of labels, we simulated the label probabilistic atlases by applying Gaussian convolution to the labels of the atlas and then normalizing the values as label probabilities.
\Zxhreffig{fig:atlas} illustrates the constructed atlas intensity image and the corresponding probabilistic atlas image of myocardium.

\subsubsection{Evaluation metrics} \label{result:evaluationmetric}
To evaluate the accuracy of segmentation results, the Dice metric and the average contour distance (ACD) between the automatic segmentation and the corresponding gold standard are calculated:
\begin{itemize}
\item  Dice:
Let $Vol_{\texttt{seg}}$ represent the volume segmented by an algorithm and $Vol_{\texttt{GD}}$ be the gold standard.
The Dice is defined related to the overlap of the two volumes,
    \begin{displaymath}
    \texttt{Dice} = \frac{2|Vol_{\texttt{seg}}\bigcap Vol_{\texttt{GD}}|} {|Vol_{\texttt{seg}}|+|Vol_{\texttt{GD}}|}
    \end{displaymath}
Here, $|Vol|$ indicates the volume size. Dice scores range from 0, meaning no overlap between the two volumes, to 1, implying perfect overlap and similarity.
\item ACD:
The ACD metric computes the average Euclidean distance between the two corresponding contours of the segmentation results.
For every point in the segmented contours, the Euclidean distance to the nearest point in the gold standard contours is calculated, and the ACD is the average of each data set.
The lower value of the ACD, the better accuracy of the segmentation.
\end{itemize}


\subsubsection{Implementations}

In the atlas-based segmentation, the atlas is registered to the target image using a hierarchical registration scheme, which is specifically designed for cardiac images and consists of three levels of transformations, i.e. affine, locally affine, and free-form deformations \citep{journal/tmi/Zhuang10}.
In the MvMM myocardial segmentation, the atlas is first registered to the bSSFP which covers the whole ventricles, and the registered atlas probabilities are propagated to the common space of the MS CMR.

The multi-component GMM and MvMM both assigns two components to the myocardium for the LGE and T2 CMR, respectively modeling the normal and abnormal tissues, and two components to the background for all the three CMR sequences; for the others, such as the left ventricular blood pool, one component is used.

The algorithms such as registration and EM iteration were implemented based on single thread using C++ code.
These tools were then wrapped up using Matlab scripts to form different segmentation pipelines.
The experiments were run on a Lenovo D30 ThinkStation which had 3.30 GHz Intel Xeon E5-2667 V2 CPU and 64 GB main memory.

\subsubsection{Performance in detail: Visual assessment}\label{result:cmr:detailvis}

\begin{figure*}[p]\center
\includegraphics[width=\textwidth]{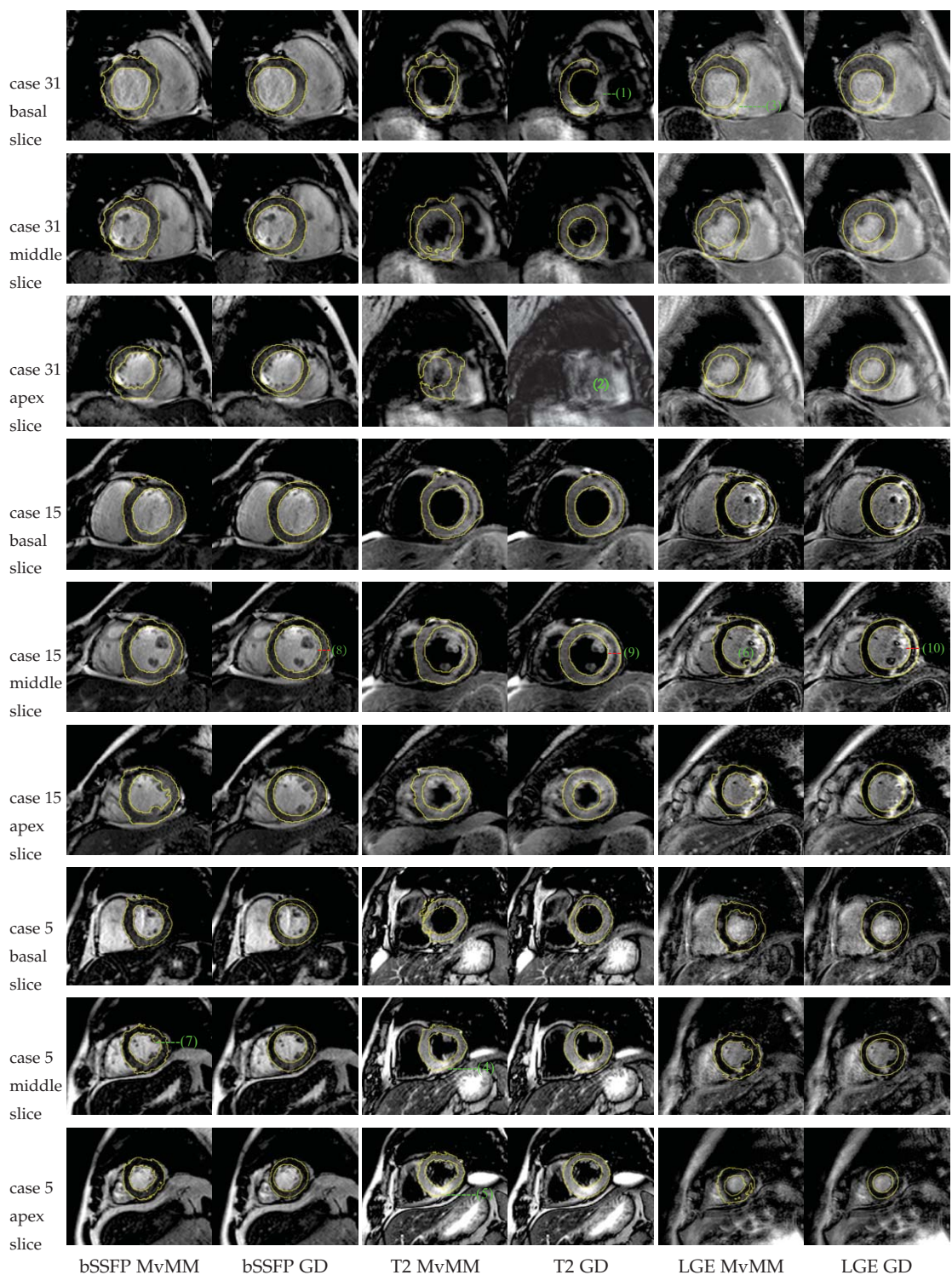}\\[-12ex]
\caption{Visual segmentation results of three \emph{median} cases in terms of Dice scores (myocardium) of the three CMR sequences. This figure presents the automatic segmentation by the MvMM and the gold standard segmentation, referred to as GD, superimposed on with the CMR images. For interpretation of color and details, readers are referred to rooming in the electronic edition of this article. }
\label{fig:resultvis}
\end{figure*}

The visual assessment provides encouraging conclusion, namely the MvMM can obtain good accuracy and robustness for the myocardial segmentation, particularly for the LGE CMR which is the most challenging task.
\Zxhreffig{fig:resultvis} visualizes three typical examples, where the basal slice, middle slice and apical slice are presented with the myocardial contours superimposed on.
These examples are selected because their Dice scores of myocardium are the median values among all the test subjects, i.e. Case 31 has the median Dice among all LGE images, Case 15 has the median Dice among all T2, and Case 5 has the median Dice among all bSSFP.

The three CMR sequences generally acquire images at different breath holds, from various positions of the heart, and with variant resolutions.
Hence, the slices from different sequences may not have corresponding slices, particularly for the T2 images.
For example, the T2 images in the three subjects in \zxhreffig{fig:resultvis} only have three slices,
the basal and apical slices of Case 31 were acquired beyond the conventional positions of the heart, and thus the gold standard segmentation from the average of the three manual delineations does not have myocardium label at some areas, as (1) and (2) in \zxhreffig{fig:resultvis} illustrate.
This leads to a low Dice score of the T2 myocardial segmentation, 0.524, for Case 31.
Similarly, T2 image in Case 5 does not have the apical slice corresponding to that of bSSFP and LGE CMR, and hence the author only presents the closest slice as the \emph{apical slice} of T2.

The LGE CMR segmentation results of Case 31 and Case 15 show that the MvMM can accurately delineate the myocardial scars, even though they appear the same as the blood pools, and no myocardial boundary is visible between them, as (3) in \zxhreffig{fig:resultvis} Case 31 indicates.
This is mainly attributed to the complimentary information of boundaries from the bSSFP and T2 images.
Also, the multi-component strategy of the proposed MvMM provides the mechanism to segment the myocardium which has inhomogeneous intensity distribution in the LGE and T2 images, such as the LGE CMR in Case 31 and Case 15, and the T2 CMR in Case 5 where the myocardial edema appears brighter than the normal tissues in the middle and apex slices, as (4) and (5) in \zxhreffig{fig:resultvis} point out.

Papillary muscles are generally regarded as part of the ventricular blood pools for consistency of many clinical applications.
However, the MvMM can sometimes include a small part of papillary muscle into the myocardium, due to the limitation of the intensity-based classification, as (6) in Case 15 and (7) in Case 5 present. This misclassification however can be corrected using some well developed post-processing algorithms, such as convex hull  \citep{conf/fimh/Shi11}.

Finally, one can observe from the three cases in \zxhreffig{fig:resultvis} that the myocardium in T2 CMR appears to be thicker than that in the LGE and bSSFP CMR sequences.  For example, the myocardium thickness equals three mark units in Case 15 (middle slice), while in the corresponding position of bSSFP and LGE CMR, the myocardium thickness appears to be around two mark units, as \zxhreffig{fig:resultvis} (8)-(10) demonstrate.
This is mainly due to the fact that the trabeculae carneae pepicardium and pericardium of the heart are more visible in T2 CMR, and also our observers tended to include them into the myocardium in the manual segmentation.
This inconsistency affects the evaluation of the combined segmentation of the MvMM, both visually and quantitatively.


\subsubsection{Performance in detail: Quantitative evaluation}\label{result:cmr:detailquan}

\begin{table*} [tb] \center
\caption{The Dice and ACD of the proposed MvMM from the three CMR sequences in details. The segmentation of T2 by GMM based on the results of MvMM are also provided for comparisons in T2(+GMM).}
\begin{tabular}{|l|ccc|cc|} \hline
\multirow{2}{*}{CMR} &  \multicolumn{3}{|c|}{Dice} & \multicolumn{2}{c|}{ACD (mm)}    \\ \cline{2-6}
                   & Endocardium & Epicardium & Myocardium  & Endocardium & Epicardium \\ \hline
LGE
&$0.866\pm0.063$ &$0.896\pm0.036$ &$0.717\pm0.076$ &$2.17\pm1.09$ &$2.16\pm0.86$ \\
T2
&$0.794\pm0.124$ &$0.908\pm0.043$ &$0.717\pm0.129$ &$2.65\pm1.50$ &$2.13\pm1.42$ \\
bSSFP
&$0.903\pm0.048$ &$0.917\pm0.027$ &$0.764\pm0.064$ &$1.82\pm1.29$ &$1.94\pm0.92$ \\[4pt]
T2{\footnotesize (+GMM)}
&$0.827\pm0.094$ &$0.878\pm0.046$ &$0.744\pm0.094$ &$2.39\pm1.60$ &$2.25\pm1.25$ \\
\hline
\end{tabular}\label{tb:resultdicesd}
\end{table*}
\begin{table*} [tb] \center
\caption{The Dice and ACD of the inter-observer (IOb) and inter-sequence (ISq) variations. }
\begin{tabular}{|l|ccc|cc|} \hline
\multirow{2}{*}{Study: CMR } &  \multicolumn{3}{|c|}{Dice} & \multicolumn{2}{c|}{ACD (mm)}    \\ \cline{2-6}
                   & Endocardium & Epicardium & Myocardium  & Endocardium & Epicardium \\ \hline
IOb: LGE
&$0.876\pm0.069$ &$0.903\pm0.041$ &$0.757\pm0.083$ &$2.08\pm0.96$ &$2.09\pm0.89$ \\
IOb: T2
&$0.894\pm0.052$ &$0.906\pm0.056$ &$0.824\pm0.069$ &$2.20\pm2.01$ &$2.60\pm2.21$ \\
IOb: bSSFP 
&$0.919\pm0.029$ &$0.927\pm0.028$ &$0.812\pm0.046$ &$1.81\pm1.04$ &$2.06\pm1.28$ \\[4pt]

ISq: T2-LGE
&$0.825\pm0.100$ &$0.898\pm0.050$ &$0.734\pm0.115$ &$2.93\pm2.05$ &$2.96\pm2.12$ \\
ISq: T2-bSSFP
&$0.778\pm0.115$ &$0.910\pm0.040$ &$0.681\pm0.106$ &$2.77\pm1.09$ &$2.25\pm1.02$ \\
ISq: LGE-bSSFP
&$0.819\pm0.084$ &$0.875\pm0.055$ &$0.660\pm0.103$ &$3.14\pm1.40$ &$2.94\pm1.45$ \\

\hline
\end{tabular}\label{tb:resultvariation}
\end{table*}

\Zxhreftb{tb:resultdicesd} presents the average Dice scores and ACD values of the proposed MvMM, which segments the three CMR sequences simultaneously.
One can see that all the three sequences have been well segmented, particularly the LGE CMR which is the most challenging and important task in cardiac segmentation.
This is mainly attributed to the combination of the complementary information, particularly from the T2 and bSSFP sequences which provide the critical boundary information for the segmentation of the LGE CMR on the scar regions.

One can notice that the Dice scores and ACD of the T2 segmentation is not as good as bSSFP, which contradicts the fact that the T2 sequence generally provides good contrast on the myocardium, and a better accuracy should be expected.
This is mainly due to the fact that the myocardium appears to be thicker in the T2 images, leading to inconsistent manual segmentation between the T2 image and the other two sequences, as we have discussed above in the \emph{visual assessment}.
The author thus further applied a GMM segmentation to the T2 images where the prior probabilities were computed from the resultant T2 segmentation of the MvMM. This segmentation, referred to as T2{\footnotesize (+GMM)}, is thus solely based on the appearance of the T2 images, and hence the evaluation results should be expected to better than the MvMM.
The quantitative results of T2{\footnotesize (+GMM)} segmentation is presented in \zxhreftb{tb:resultdicesd}, where one can find that the myocardium Dice score has been indeed increased, mainly thanks to the improvement of the endocardium. The epicardium is however worsen by T2{\footnotesize (+GMM)} due to the indistinctness of boundaries and lacking of assistance from the other two sequences.

Finally, the mean runtime of the MvMM for registering and segmenting the three CMR sequences is $9.17\pm3.95$ min.

\subsubsection{Variation study: Inter-observer and inter-sequence}
\Zxhreftb{tb:resultvariation} gives the \emph{inter-observer} variations from the three manual delineations and the \emph{inter-sequence} variations among the three sequences.
In computing the inter-sequence variations, the manual segmentation of the high-resolution images was transformed onto the image space with lower resolution, after a global affine registration to correct the global misalignment due to body motions, and the Dice scores and ACD were solely computed on the overlapped regions of the two CMR sequences.
In this study, both the LGE and bSSFP CMR were transformed onto the T2 image space and the LGE images were transformed onto the bSSFP image space.

The inter-sequence inconsistency is evident, mainly due to two reasons.
One is that the intra-image motion shift between slices of the images were not corrected, though the global motion shift was corrected using affine registration. This inter-slice shift of two CMR sequences can be very different. For example, one of the first sequence can be shifted to the left side of the whole heart, while the corresponding slice in the other sequence can be shifted entirely to the opposite direction, the right side.
The other reason is that there can exist large deformation of the heart between two scans, while in this study only affine transformation is assumed for correcting inter-sequence motions, leading to big difference of the segmentation results from two CMR sequences.

\subsection{LGE CMR Segmentation Using Different Schemes} \label{exp:alternative}

\begin{table*} [tb] \center
\caption{This table provides the Dice scores of the LGE CMR segmentation by the four separate segmentation methods and four combined segmentation schemes. Please refer to the text for details of the methods.}
\label{tb:resultcompare}
\begin{tabular}{ l cccc }
\hline\hline
Separate Seg     & Atlas          & GMM            & Atlas+bSSFP    & GMM+bSSFP   \\ \hline
Endocardium  &$0.7335\pm0.1661$ &$0.7321\pm0.1549$ &$0.8307\pm0.0740$ &$0.8357\pm0.0713$   \\
Epicardium   &$0.8160\pm0.1149$ &$0.7972\pm0.1103$ &$0.8842\pm0.0422$ &$0.8710\pm0.0557$  \\
Myocardium   &$0.4950\pm0.1959$ &$0.5163\pm0.1989$ &$0.6203\pm0.1117$ &$0.6352\pm0.1202$   \\  \hline\hline
Combined Seg     & Mvmm$^\ominus$ & Mvmm$^\ominus+$FFD & Mvmm$^\ominus+$SC   & MvMM \\ \hline
Endocardium  &$0.8552\pm0.0649$ &$0.8569\pm0.0640$ &$0.8654\pm0.0638$ &$0.8657\pm0.0633$ \\
Epicardium   &$0.8919\pm0.0384$ &$0.8912\pm0.0396$ &$0.8958\pm0.0356$ &$0.8958\pm0.0356$ \\
Myocardium   &$0.6971\pm0.0864$ &$0.7008\pm0.0851$ &$0.7168\pm0.0762$ &$0.7169\pm0.0760$ \\  \hline\hline
\end{tabular}
\end{table*}

\begin{figure*}[tb]\center
\includegraphics[width=\textwidth]{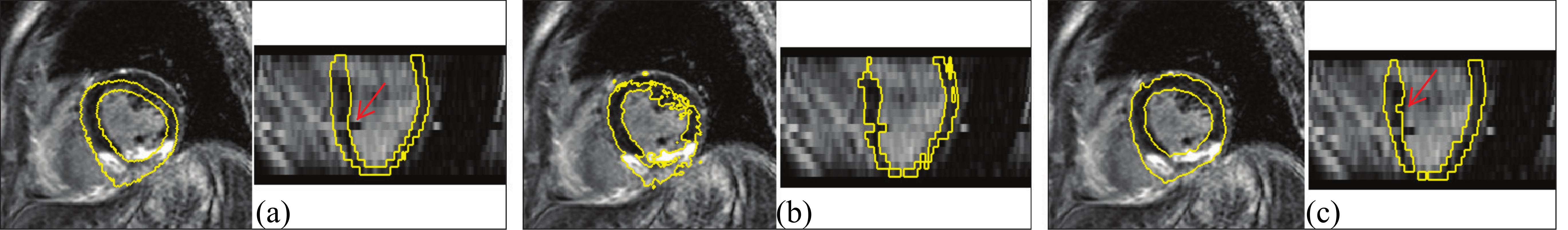}
\caption{The LGE CMR segmentation results using the atlas-based segmentation (a), GMM segmentation (b), and the Mvmm$^\ominus$ without registration correction (c). The segmentation result using the proposed MvMM is presented in \zxhreffig{fig:misalignment}. }
\label{fig:resultcompare}\end{figure*}

This section first compares four conventional methods which perform the segmentation of each CMR sequence separately, including,
\begin{enumerate}
\item[(1)] the atlas-based segmentation using the initial result computed directly from the registered probabilistic atlases, referred to as \emph{Atlas};
\item[(2)] the conventional univariate GMM segmentation initialized by the propagated probabilities from the atlas, referred to as \emph{GMM};
\item[(3)] the atlas-based segmentation where the bSSFP image of the same subject is used to assist the atlas-to-target registration, referred to as \emph{Atlas+bSSFP};
\item[(4)] the GMM segmentation initialized from the result of Atlas+bSSFP, referred to as \emph{GMM+bSSFP}.
\end{enumerate}
Then, four MvMM segmentation methods, implemented with different registration schemes, are studied,
\begin{enumerate}
\item[(1)] the MvMM scheme without any registration correction, referred to as Mvmm$^\ominus$;
\item[(2)] the Mvmm$^\ominus$ plus FFD registration for atlas correction, referred to as Mvmm$^\ominus+$FFD;
\item[(3)] the the Mvmm$^\ominus$ plus shift correction for the motion shifted slices, referred to as Mvmm$^\ominus+$SC;
\item[(4)] the proposed MvMM with both FFD and SC registration correction, referred to as MvMM.
\end{enumerate}

\subsubsection{Comparisons with conventional methods}

The Dice scores of the four separate segmentation for the LGE CMR sequence are presented in \Zxhreftb{tb:resultcompare}.
Compared with the proposed MvMM which combines and segments the three sequences simultaneously, all the four separate segmentation schemes obtained evidently and significantly worse myocardium Dice scores ($p\!\!<\!\!0.01$).
The advantage of including complimentary information from other sequences can be confirmed by comparing the results between the four separate segmentation methods:
with the assistance of the bSSFP sequence, both Atlas+bSSFP and GMM+bSSFP achieved significantly better myocardium Dice scores, $p\!\!<\!\!0.01$, than the conventional Atlas and GMM methods.

\Zxhreffig{fig:resultcompare} displays the segmentation results using the Atlas (a) and GMM (b). One can see that the registration could not correctly align the atlas and the target image in the motion shifted slice, as the red arrow point out in \Zxhreffig{fig:resultcompare} (a). The GMM segmentation could correct part of the misclassification, but the delineation was poor in the scar area, where no boundary constraint from other images was applied, and the area which was not well initialized by the atlas, as \Zxhreffig{fig:resultcompare} (b) shows.
By contrast, the proposed Mvmm produced a much better myocardial segmentation on the LGE CMR with the guidance and constraints from the other two sequences, as \zxhreffig{fig:misalignment} (c) shows.

\subsubsection{Comparisons using different registration schemes} \label{result:misalignment}

The Dice scores of the four combined segmentation for the LGE CMR sequence are presented in \Zxhreftb{tb:resultcompare}.
Here, two types of registration correction are studied separately, i.e. (1) the FFD registration for atlas correction, (2) the affine registration for correcting inter-slice motion shift.
One can find from the results that the registration improved the segmentation performance of Mvmm$^\ominus$. The improvement on the myocardium Dice scores was evident and statistically significant ($p\!\!=\!\!0.005$) for Mvmm$^\ominus+$SC, and was significant ($p\!\!=\!\!0.008$) but marginal for Mvmm$^\ominus+$FFD.
Also, the difference between the Dice scores of Mvmm$^\ominus+$SC and MvMM is trivial and non-significant ($p\!\!=\!\!0.665$), indicating that the initial atlas-to-target registration has performed well and the shift correction should be the main concern of the registration correction.
\Zxhreffig{fig:resultcompare} (c) further illustrates that Mvmm$^\ominus$ can accurately include the scar into the myocardium, but still erroneously delineates the shifted slices, including the slice pointed out by the red arrow in (a).
\Zxhreffig{fig:resultcompare} also shows that  Mvmm$^\ominus$ misclassifies one slice, pointed out by the red arrow in (c).
This is due to the shifted slice in the bSSFP image, which corresponds to the position pointed out by the red arrow in \zxhreffig{fig:misalignment} (a).

\subsection{Study MvMM Using Simulated Brain MR} \label{exp:brain}

\begin{figure*}[tb]\center
\framebox{\includegraphics[width= 0.8\textwidth]{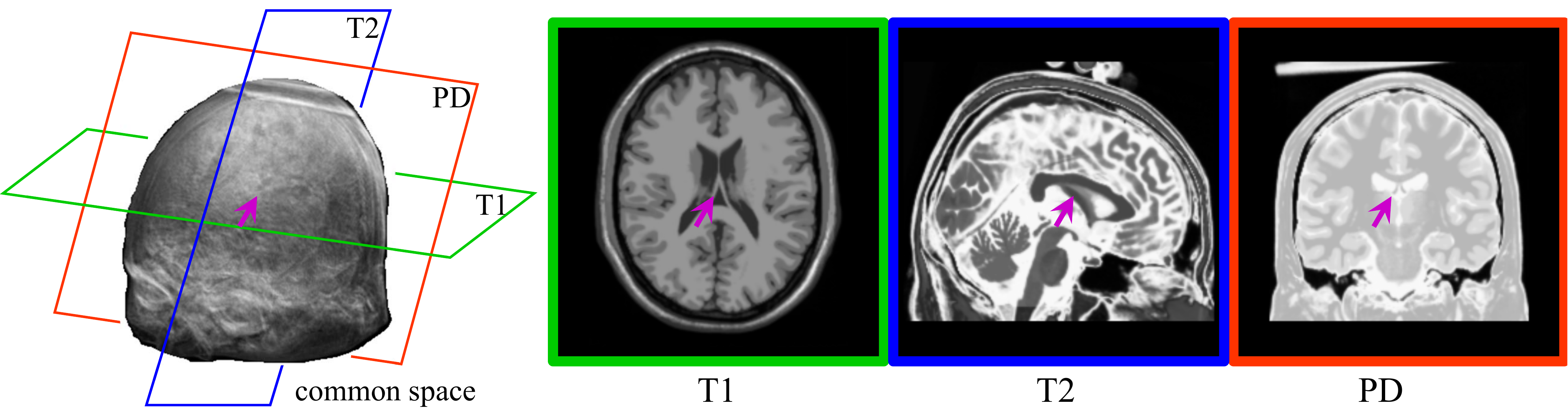}}
   \caption{Illustration of multivariate brain images in the common space, and the slices from the three brain MR sequences.}
\label{fig:commonspacebrain}\end{figure*}

\subsubsection{Data and experimental setup}
The three sequences of brain MR images, including the T1-weighted, T2-weighted and photon density (PD), are used
to study the performance of the MvMM with different number of images and variant coverage of ROI.

The three original brain MR images and the corresponding segmentation were downloaded from the Brain Web, with image size $181\!\times\!217\!\times\!181$ and voxel size $1\!\times\!1\!\times\!1$ mm.
\Zxhreffig{fig:commonspacebrain} provides three slices of the three brain MR sequences from different positions of the common space. 
The probabilistic atlases of the interested labels of the brain were generated by applying a Gaussian smoothing operation on the extracted labels, followed a normalization process as the probabilities of labels on each pixel.

To simulate fifty segmentation cases for each experiment, this study used a set of deformation fields to deform the original brain MR images into new ones, which was slightly misaligned to the probabilistic atlases.
Each of the deformation fields was generated using a FFD with $20\!\!\times\!\!20\!\!\times\!\!20$ mm mesh spacing, which had each control point randomly displaced following a normal distribution of 0 mean and 2 mm standard deviation.
For each case, the three brain MR sequences were deformed using different FFD transformations, and hence the MS brain MR images from one subject had mis-registration to each other.

\begin{figure*}[tb]\center \begin{tabular}{@{}c@{}c@{}}
   \includegraphics[width= 0.48\textwidth]{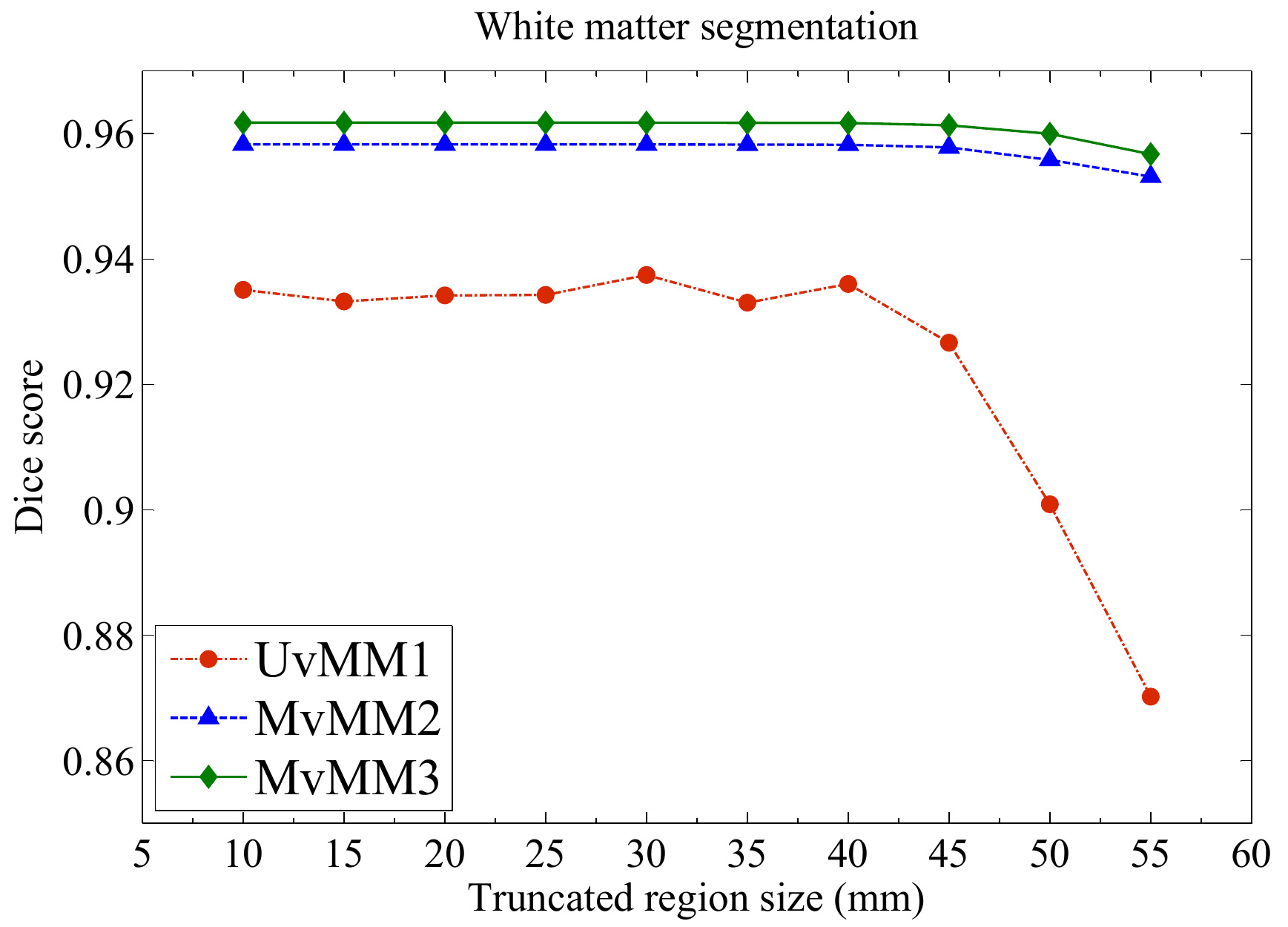} &
   \includegraphics[width= 0.48\textwidth]{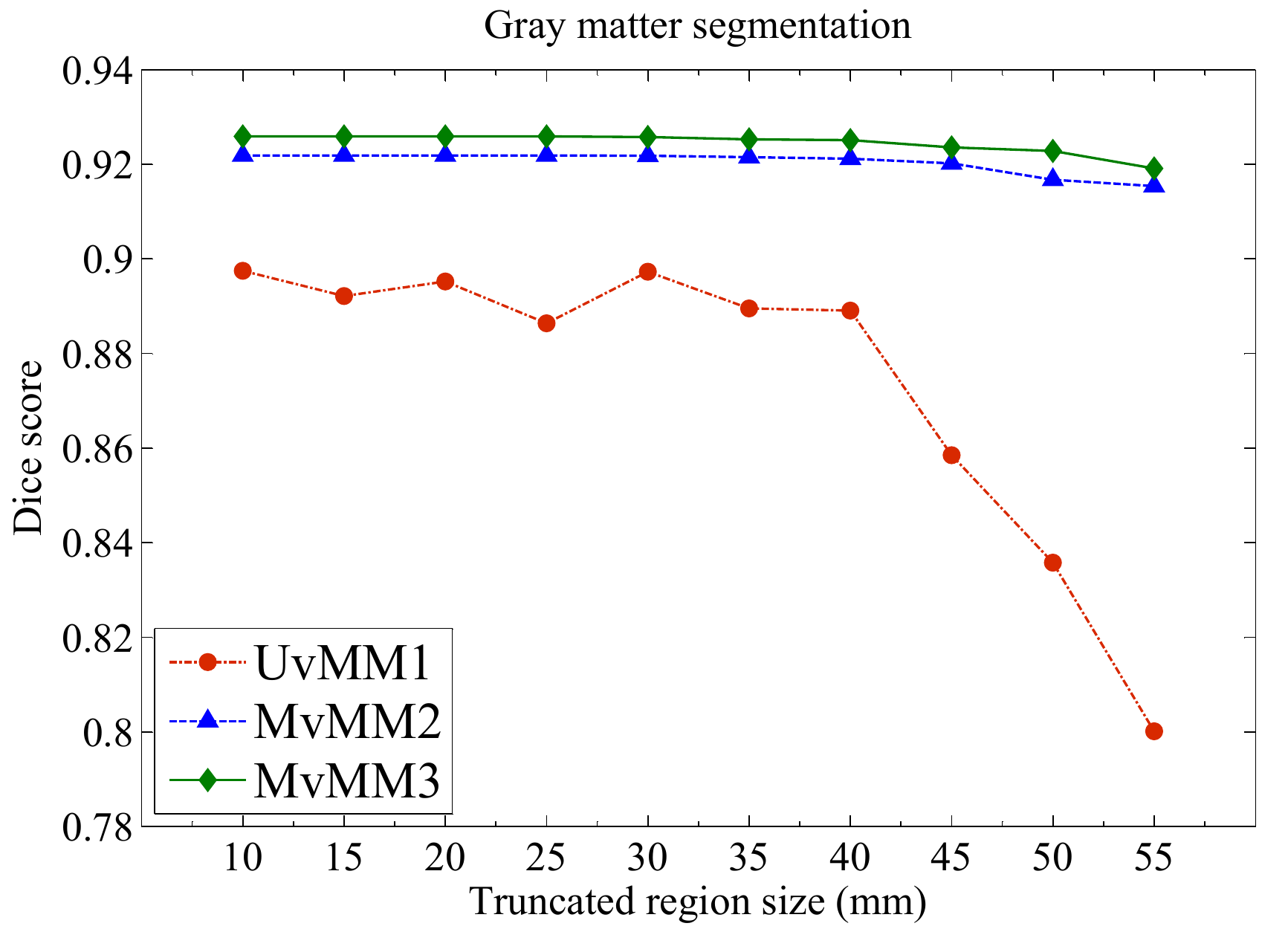}   \\ (a)&(b)\end{tabular}
   \caption{The Dice scores of the brain white matter and gray matter segmentation combining multi-sequence brain MR.}
\label{fig:bmrresult}\end{figure*}

The Dice scores of the segmentation results are presented in \zxhreffig{fig:bmrresult}, including the results of (1) Univariate Mixture Model using single image, referred to as UvMM1,
(2) MvMM using two images where the MVI is a bivariate vector, referred to as MvMM2,
and (3) MvMM using three images where the MVI is a tri-variate variable, referred to as MvMM3.

\subsubsection{MVIs with different dimensions}
One can see from \Zxhreffig{fig:bmrresult} that the segmentation performance, indicated by the Dice scores, can been improved with respect to the increased dimensions of the variate, i.e. the number of images to form the MVIs. In general, the improvement from the univariate to bivariate is more evident than that from the bivariate to tri-variate, indicating the performance tends to converge after certain number of images are combined for simultaneous segmentation.

\subsubsection{Hetero-Coverage MVIs}
In this experiment, the author manually reduced the ROI of an image by deleting a number of slices to truncate the image volume, to simulate the HC-MVIs. Given an image, this truncation was done randomly at one end of the three dimensions, for example one image may have the first $l$ slices at x-dimension truncated, while the other may have the end $k$ slices at the z-dimension truncated.
We indicate the sizes of the truncated regions using sum of the thickness of all the deleted slices.

\Zxhreffig{fig:bmrresult} displays the Dice scores of the segmented white matter and gray matter using the three segmentation schemes.
The Dice scores of MvMM2 and MvMM3 are much higher than that of UvMM1.
Particularly, the two MvMM schemes still achieved good performance even when the images were randomly truncated more than 40 mm stack of slices, which compares differently with UvMM1.
This confirms the advantage of MvMM combining multi-source images for simultaneous segmentation, namely the combined segmentation improves the robustness against some images which do not fully cover the ROI.

\section{Conclusion and Discussion}\label{conclude}

The author has presented a new method, i.e. MvMM, for simultaneous segmentation of multi-source images, particularly for the myocardial segmentation combining the complementary information from multi-sequence (MS) CMR.
In this method, the MS CMR of the same subject are aligned to a common space and the segmentation of them is performed simultaneously.
To correct the misalignments of slices due to the motion shift and the mis-registration of atlases, the MvMM and LL are formulated with transformations and the ICM approach is employed to update the two groups of parameters, namely the MvMM (segmentation) parameters are optimized using the EM algorithm and the transformation (registration) variables are updated using the gradient ascent method.

In the experiments, the author first investigated the performance of the proposed MvMM for myocardial segmentation of CMR sequences.
The average Dice scores for LGE CMR segmentation were respectively 0.866 (endocardium), 0.896 (epicardium), and 0.717 (myocardium); and the ACD for the endocardium and epicardium were 2.17 mm and 2.16 mm respectively.
These results are promising, by comparing them with the corresponding values from the inter-observer variations, which were
0.876 (endocardium), 0.903 (epicardium), 0.757 (myocardium), 2.08 mm (ACD of endocardium) and 2.09 mm (ACD of epicardium).
Generally, the bSSFP can obtain good accuracy, thanks to the high quality of images which also contributes to reducing inconsistency of manual segmentation (inter-observer variation).
The Dice scores and ACD of T2 CMR were not as good as that of bSSFP in the experiments, due to the inconsistent inter-sequence variations.
A simple follow-up GMM segmentation based on the results of MvMM had shown to improve the accuracy of the T2 CMR segmentation.

To compare with the conventional methods, the author included the atlas-based segmentation and the conventional GMM method for separately segmentation on the LGE CMR. Both the Dices scores of these two methods were statistically and evidently worse than the proposed MvMM ($p\!\!<\!\!0.01$).
The author further included the bSSFP CMR for the initial registration of the Atlas and GMM segmentation. The usage of bSSFP significantly improved the Dice scores of the segmentation ($p\!\!<\!\!0.01$), but the separate segmentation results, either the Atlas+bSSFP or GMM+bSSFP, were still significantly worse than the MvMM ($p\!\!<\!\!0.01$).

In Section~\ref{exp:alternative}, the author further investigated the MvMM without registration, MvMM with nonrigid atlas-to-target registration, and MvMM with affine registration for correcting motion shift of slices.
The registration in each step could generally improve the segmentation performance of the MvMM, but the improvement from the atlas-to-target registration was not as evident as the correction of motion shift.

In Section~\ref{exp:brain}, the author employed brain MR to study the gain of segmentation performance by using MVIs with higher dimensions and investigate the performance of MvMM in HC-MVIs. The MvMM schemes were evidently better than the UvMM1 segmentation, but the performance of MvMM2 was comparable to MvMM3, indicating a convergence of segmentation performance by the high-dimensional MvMM after using bivariate images (two-source images).
It is worth mentioning that the HC-MVIs experiment showed that the combined MvMM segmentation was much more robust than the separate UvMM1 segmentation in the applications when some images had less coverage of the ROI.

In the literature, there have been limited works focusing on the fully automatic LGE CMR segmentation, which is however an important prerequisite in a number of clinical applications of cardiology.
Two conference papers \cite{conf/miccai/Dikici04, conf/isbi/Ciofolo08} have reported about 2.25 mm ACD (42 slices) and 2.1 mm ACD (27 subects) for LGE CMR myocardium segmentation results, respectively.
The authors in \cite{journal/mia/Wei13} implemented manual interaction to correct the mis-registration when the automatic translational registration failed on certain slices, and the average Dice score of myocardium segmentation was 0.82 on 21 subjects.
In these three works, the segmentation of LGE CMR was propagated from the segmentation results of corresponding bSSFP image of the same subject at the same cardiac phase.
The bSSFP segmentation provides strong prior for the segmentation of LGE CMR. However, the application of the shape from bSSFP is not fully automatic, nor integrated within a unified framework for simultaneous and combined segmentation.
In the preliminary conference paper \citep{conf/miccai/Zhuang16}, the author tested the MvMM on 18 selected data sets, which had good image quality and small inter-sequence variation.
The average Dice score of myocardium segmentation could reach up to 0.74.
It should be noted that an objective inter-study comparison can be difficult, since the data sets and implementation can vary greatly across different studies.

There are three limitation of this work.
First, the adoption of T2 CMR may be a sub-optimal choice, as T2 CMR generally only has a few slices, such as three in many of our clinical data; and it also images myocardium which appears to be thicker than the LGE and bSSFP CMR, leading to inconsistent manual delineations for the validation of myocardial segmentation.
Second, there can exist nonrigid misalignment between differen sequences, for example due to the respiratory motions during CMR acquisitions. Therefore, efficient methods for modeling such deformations should be considered in the future work.
Finally, notably the myocardial segmentation task from MS CMR is a selected example for demonstrating the application of the proposed MvMM, which nevertheless is generally applicable to the segmentation and classification tasks where multi-source images are available and can be combined.
For example, in Section~\ref{exp:brain} the MvMM was applied to the segmentation of MS brain MR.
Therefore, the future work will be to extend the MvMM and multivariate image analysis framework to other applications where the multi-modality multi-source imaging data are commonly seen in the clinics.


\bibliographystyle{IEEEtran}

\begin{thebibliography}{10}
\providecommand{\url}[1]{#1}
\csname url@samestyle\endcsname
\providecommand{\newblock}{\relax}
\providecommand{\bibinfo}[2]{#2}
\providecommand{\BIBentrySTDinterwordspacing}{\spaceskip=0pt\relax}
\providecommand{\BIBentryALTinterwordstretchfactor}{4}
\providecommand{\BIBentryALTinterwordspacing}{\spaceskip=\fontdimen2\font plus
\BIBentryALTinterwordstretchfactor\fontdimen3\font minus
  \fontdimen4\font\relax}
\providecommand{\BIBforeignlanguage}[2]{{%
\expandafter\ifx\csname l@#1\endcsname\relax
\typeout{** WARNING: IEEEtran.bst: No hyphenation pattern has been}%
\typeout{** loaded for the language `#1'. Using the pattern for}%
\typeout{** the default language instead.}%
\else
\language=\csname l@#1\endcsname
\fi
#2}}
\providecommand{\BIBdecl}{\relax}
\BIBdecl

\bibitem{journal/arbe/PhamXP00}
D.~L. Pham, C.~Xu, and J.~L. Prince, ``Current methods in medical image
  segmentation,'' \emph{Annual Review of Biomedical Engineering}, vol.~02, pp.
  315--337, 2000.

\bibitem{journal/tmi/NobleB06}
J.~A. Noble and D.~Boukerroui, ``Ultrasound image segmentation: {A} survey,''
  \emph{IEEE Transaction on Medical Imaging}, vol.~25, no.~8, pp. 987--1010,
  Aug. 2006.

\bibitem{journal/mia/Petitjean11}
C.~Petitjean and J.~N. Dacher, ``{{A} review of segmentation methods in short
  axis cardiac {M}{R} images},'' \emph{Med Image Anal}, vol.~15, no.~2, pp.
  169--184, Apr 2011.

\bibitem{journal/circ/Kim09}
R.~J. Kim, D.~S. Fieno, T.~B. Parrish, K.~Harris, E.-L. Chen, O.~Simonetti,
  J.~Bundy, J.~P. Finn, F.~J. Klocke, and R.~M. Judd, ``Relationship of mri
  delayed contrast enhancement to irreversible injury, infarct age, and
  contractile function,'' \emph{Circulation}, vol. 100, no.~19, pp. 1992--2002,
  1999.

\bibitem{journal/jacc/Kim09}
H.~W. Kim, A.~Farzaneh-Far, and R.~J. Kim, ``Cardiovascular magnetic resonance
  in patients with myocardial infarctioncurrent and emerging applications,''
  \emph{Journal of the American College of Cardiology}, vol.~55, no.~1, pp.
  1--16, 2009.

\bibitem{journal/ijci/Kolipaka05}
A.~Kolipaka, G.~P. Chatzimavroudis, R.~D. White, T.~P. O¡¯Donnell, and R.~M.
  Setser, ``Segmentation of non-viable myocardium in delayed enhancement
  magnetic resonance images,'' \emph{The international journal of
  cardiovascular imaging}, vol.~21, no. 2-3, pp. 303--311, 2005.

\bibitem{journal/jacc/Flett11}
A.~S. Flett, J.~Hasleton, C.~Cook, D.~Hausenloy, G.~Quarta, C.~Ariti,
  V.~Muthurangu, and J.~C. Moon, ``Evaluation of techniques for the
  quantification of myocardial scar of differing etiology using cardiac
  magnetic resonance,'' \emph{JACC: cardiovascular imaging}, vol.~4, no.~2, pp.
  150--156, 2011.

\bibitem{journal/tmi/LeemputMS99}
K.~V. Leemput, F.~Maes, D.~Vandermeulen, and P.~Suetens, ``Automated
  model-based tissue classification of {MR} images of the brain,'' \emph{IEEE
  Transactions on Medical Imaging}, vol.~18, no.~10, pp. 897--908, 1999.

\bibitem{journal/mia/ValdesSR04}
M.~Lorenzo-Valdes, G.~I. Sanchez-Ortiz, A.~G. Elkington, R.~Mohiaddin, and
  D.~Rueckert, ``Segmentation of 4{D} cardiac {MR} images using a probabilistic
  atlas and the {EM} algorithm,'' \emph{Medical Image Analysis}, vol.~8, pp.
  255--265, 2004.

\bibitem{conf/embc/Berbari09}
R.~Berbari, N.~Kachenoura, F.~Frouin, A.~Herment, E.~Mousseaux, and I.~Bloch,
  ``An automated quantification of the transmural myocardial infarct extent
  using cardiac de-mr images,'' in \emph{Conf Proc IEEE Eng Med Biol Soc},
  vol.~1, 2009, pp. 4403--4406.

\bibitem{conf/miccai/Dikici04}
E.~Dikici, T.~O'Donnell, R.~Setser, and R.~White, ``Quantification of delayed
  enhancement {MR} images,'' in \emph{Medical Image Computing and
  Computer-Assisted Intervention}, vol. 3216, 2004, pp. 250--257.

\bibitem{conf/isbi/Xu13}
R.~S. Xu, P.~Athavale, Y.~Lu, P.~Radau, and G.~A. Wright, ``Myocardial
  segmentation in late-enhancement mr images via registration and propagation
  of cine contours,'' in \emph{2013 IEEE 10th International Symposium on
  Biomedical Imaging}.\hskip 1em plus 0.5em minus 0.4em\relax IEEE, 2013, pp.
  856--859.

\bibitem{conf/isbi/Ciofolo08}
C.~Ciofolo, M.~Fradkin, B.~Mory, G.~Hautvast, and M.~Breeuwer, ``Automatic
  myocardium segmentation in late-enhancement mri,'' in \emph{2008 5th IEEE
  International Symposium on Biomedical Imaging: From Nano to Macro}.\hskip 1em
  plus 0.5em minus 0.4em\relax IEEE, 2008, pp. 225--228.

\bibitem{journal/tmi/Rajchl14}
M.~Rajchl, J.~Yuan, J.~White, E.~Ukwatta, J.~Stirrat, C.~Nambakhsh, F.~Li, and
  T.~Peters, ``Interactive hierarchical-flow segmentation of scar tissue from
  late-enhancement cardiac {MR} images,'' \emph{IEEE Transactions on Medical
  Imaging}, vol.~33, pp. 159--172, 2014.

\bibitem{journal/mia/Wei13}
D.~Wei, Y.~Sun, S.-H. Ong, P.~Chai, L.~L. Teo, and A.~Low, ``Three-dimensional
  segmentation of the left ventricle in late gadolinium enhanced mr images of
  chronic infarction combining long- and short-axis information,''
  \emph{Medical Image Analysis}, vol.~17, pp. 685--697, 2013.

\bibitem{geladi1996multivariate}
P.~Geladi and H.~F. Grahn, \emph{Multivariate image analysis}.\hskip 1em plus
  0.5em minus 0.4em\relax Wiley Online Library, 1996.

\bibitem{prats2011multivariate}
J.~M. Prats-Montalb{\'a}n, A.~De~Juan, and A.~Ferrer, ``Multivariate image
  analysis: a review with applications,'' \emph{Chemometrics and Intelligent
  Laboratory Systems}, vol. 107, no.~1, pp. 1--23, 2011.

\bibitem{bharati1998multivariate}
M.~Bharati and J.~MacGregor, ``Multivariate image analysis for real-time
  process monitoring and control,'' \emph{Industrial \& Engineering Chemistry
  Research}, vol.~37, no.~12, pp. 4715--4724, 1998.

\bibitem{conf/fimh/Shi11}
W.~Shi, X.~Zhuang, H.~Wang, S.~Duckett, D.~Oregan, P.~Edwards, S.~Ourselin, and
  D.~Rueckert, ``Automatic segmentation of different pathologies from cardiac
  cine {MRI} using registration and multiple component {EM} estimation,'' in
  \emph{Functional Imaging and Modeling of the Heart}, 2011, pp. 163--170.

\bibitem{journal/ni/Ashburner05}
J.~Ashburner and K.~J. Friston, ``Unified segmentation,'' \emph{NeuroImage},
  vol.~26, no.~3, pp. 839--851, 2005.

\bibitem{journal/tmi/Zhuang10}
X.~Zhuang, K.~Rhode, R.~Razavi, D.~J. Hawkes, and S.~Ourselin, ``A
  registration-based propagation framework for automatic whole heart
  segmentation of cardiac {MRI},'' \emph{IEEE Transactions on Medical Imaging},
  vol.~29, no.~9, pp. 1612--1625, 2010.

\bibitem{journal/mia/Zhuang16}
X.~Zhuang and J.~Shen, ``Multi-scale patch and multi-modality atlases for whole
  heart segmentation of {MRI},'' \emph{Medical Image Analysis}, vol.~31, pp.
  77--87, 2016.

\bibitem{conf/spie/Lee00}
S.-J. Lee, ``Accelerated coordinate descent methods for bayesian reconstruction
  using ordered subsets of projection data,'' in \emph{Proc. SPIE 4121,
  Mathematical Modeling, Estimation, and Imaging}, 2000.

\bibitem{journal/tmi/Rueckert99}
D.~Rueckert, L.~I. Sonoda, C.~Hayes, D.~L.~G. Hill, M.~O. Leach, and D.~J.
  Hawkes, ``Nonrigid registration using free-form deformations: {A}pplication
  to breast {MR} images,'' \emph{IEEE Transactions on Medical Imaging},
  vol.~18, pp. 712--721, 1999.

\bibitem{journal/tmi/Zhuang11}
X.~Zhuang, S.~Arridge, D.~J. Hawkes, and S.~Ourselin, ``A nonrigid registration
  framework using spatially encoded mutual information and free-form
  deformations,'' \emph{IEEE Transactions on Medical Imaging}, vol.~30, no.~10,
  pp. 1819--1828, 2011.

\bibitem{cocosco1997brainweb}
C.~A. Cocosco, V.~Kollokian, R.~K.-S. Kwan, G.~B. Pike, and A.~C. Evans,
  ``Brainweb: Online interface to a 3d mri simulated brain database,'' in
  \emph{NeuroImage, Proceedings of 3-rd International Conference on Functional
  Mapping of the Human Brain}, vol.~5, no.~4, p. S425.

\bibitem{itksnap}
P.~A. Yushkevich, J.~Piven, H.~Cody~Hazlett, R.~Gimpel~Smith, S.~Ho, J.~C. Gee,
  and G.~Gerig, ``User-guided {3D} active contour segmentation of anatomical
  structures: Significantly improved efficiency and reliability,''
  \emph{Neuroimage}, vol.~31, no.~3, pp. 1116--1128, 2006.

\bibitem{conf/miccai/Zhuang16}
X.~Zhuang, ``Multivariate mixture model for cardiac segmentation from
  multi-sequence mri,'' in \emph{International Conference on Medical Image
  Computing and Computer-Assisted Intervention}.\hskip 1em plus 0.5em minus
  0.4em\relax Springer, 2016, pp. 581--588.

\end{thebibliography}

%








\end{document}